\documentclass[journal,12pt,onecolumn,draftclsnofoot]{IEEEtran}
\usepackage{ifpdf}

\ifpdf
\else
\fi
\usepackage{cite}
\usepackage{subfig}
\usepackage{epsfig}
\usepackage{epstopdf}
\usepackage{graphicx}
\usepackage{multicol,lipsum}
\usepackage{subfiles}
\usepackage{algorithm}
\usepackage[dvipsnames]{xcolor}
\usepackage[noend]{algpseudocode}

\usepackage{amsmath,amssymb}
\usepackage[inline]{enumitem}

\usepackage{array}

\usepackage{fixltx2e}
\begin{document}
\title{AgRegNet: A Deep Regression Network for Flower and Fruit Density Estimation, Localization, and Counting in Orchards}

\author{Uddhav~Bhattarai,
        Santosh~Bhusal,
        Qin~Zhang,
        and~Manoj~Karkee
\thanks{The authors are with Department of Biological Systems Engineering, Center for Precision and Automated Agricultural Systems, Washington State University, Prosser, WA, 99350 USA (email: uddhav.bhattarai@wsu.edu, santosh.bhusal@wsu.edu, qinzhang@wsu.edu, manoj.karkee@wsu.edu)}}


\maketitle

\begin{abstract}

One of the major challenges for the agricultural industry today is the uncertainty in manual labor availability and the associated cost. Automated flower and fruit density estimation, localization, and counting could help streamline harvesting, yield estimation, and crop-load management strategies such as flower and fruitlet thinning. This article proposes a deep regression-based network, AgRegNet, to estimate density, count, and location of flower and fruit in tree fruit canopies without explicit object detection or polygon annotation. Inspired by popular U-Net architecture, AgRegNet is a U-shaped network with an encoder-to-decoder skip connection and modified ConvNeXt-T as an encoder feature extractor. AgRegNet can be trained based on information from point annotation and leverages segmentation information and attention modules (spatial and channel) to highlight relevant flower and fruit features while suppressing non-relevant background features. Experimental evaluation in apple flower and fruit canopy images under an unstructured orchard environment showed that AgRegNet achieved promising accuracy as measured by Structural Similarity Index (SSIM), percentage Mean Absolute Error (pMAE) and mean Average Precision (mAP) to estimate flower and fruit density, count, and centroid location, respectively. Specifically, the SSIM, pMAE, and mAP values for flower images were 0.938, 13.7\%, and 0.81, respectively. For fruit images, the corresponding values were 0.910, 5.6\%, and 0.93. Since the proposed approach relies on information from point annotation, it is suitable for sparsely and densely located objects. This simplified technique will be highly applicable for growers to accurately estimate yields and decide on optimal chemical and mechanical flower thinning practices.

\end{abstract}


%
\IEEEpeerreviewmaketitle

\section{Introduction}

\IEEEPARstart{T}{he} United States tree fruit production industry is largely dependent on the human labor force for field operations such as training, harvesting, pruning, and flower and fruitlet thinning. Automation in tree fruit crops can improve the efficiency, quality, and sustainability of fruit production while reducing labor costs and addressing the labor shortage challenges.
Crop-load management approaches such as flower and fruit thinning, and harvesting require information such as flower and fruit distribution, count, and location over the different growth stages. Such information during flowering and green fruitlet development stages would be essential to leverage selective/targeted mechanical and chemical thinning approaches to maximize fruit quality and yield \cite{farjon2019detection,wang2020side}. Counting matured fruit during harvest season assists in pre-and post-harvest management strategies, including estimating the labor force, harvesting equipment, transport logistics, post-harvest processing, packing, inventory management, and sales planning. Fruit location and distribution estimation help optimize the distribution of the labor force and harvesting containers. Currently, growers manually inspect flower and fruit distribution in sample location, followed by extrapolation to the entire orchard, which is prone to error, labor-intensive, challenging to scale, and expensive \cite{gongal2016apple}.

\begin{figure}[!htb]
    
    \centering
    \includegraphics[width=0.4\textwidth]{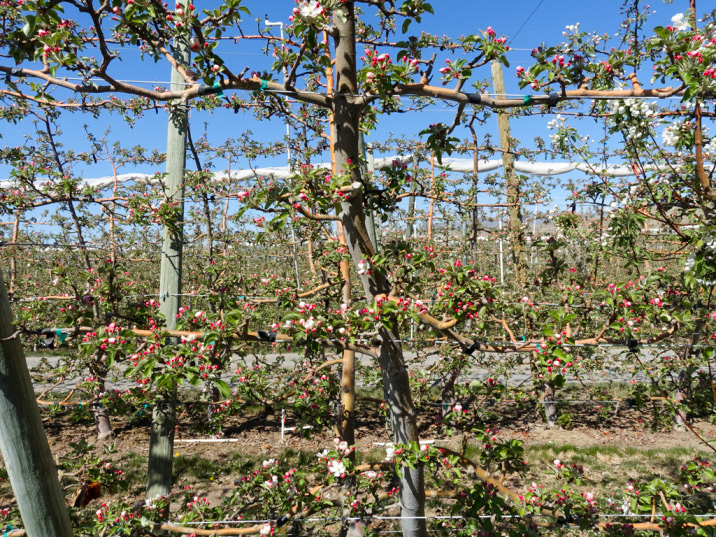}
    \includegraphics[width=0.4\textwidth]{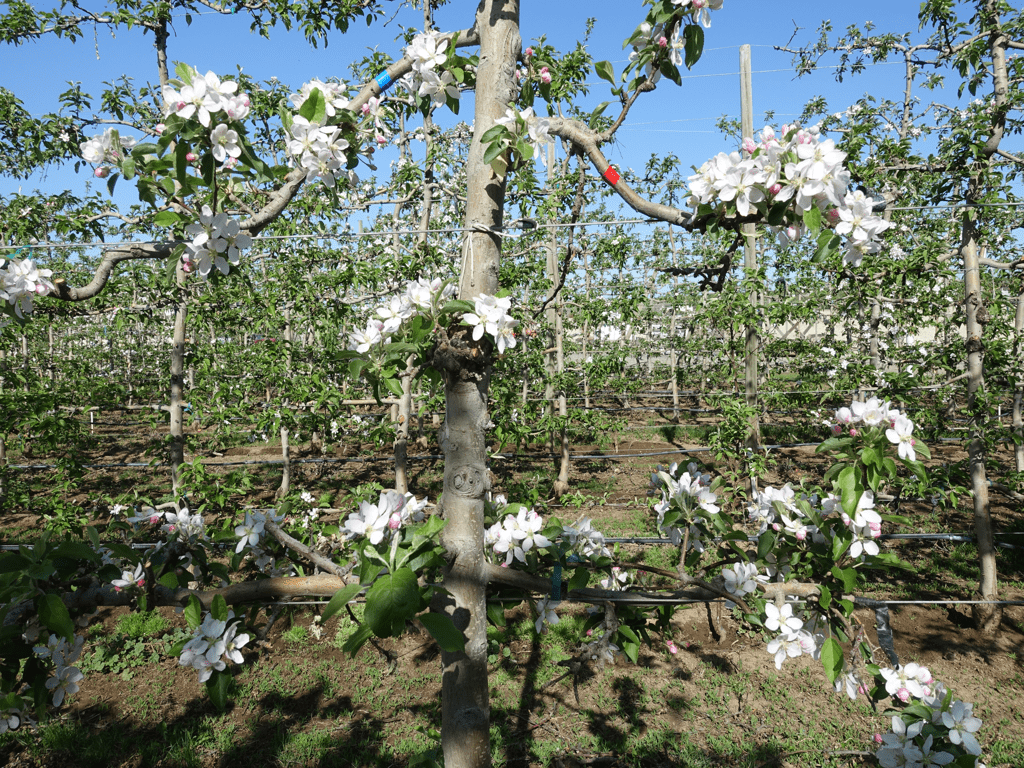} \\
    \vspace{0.8mm}
    \includegraphics[width=0.4\textwidth]{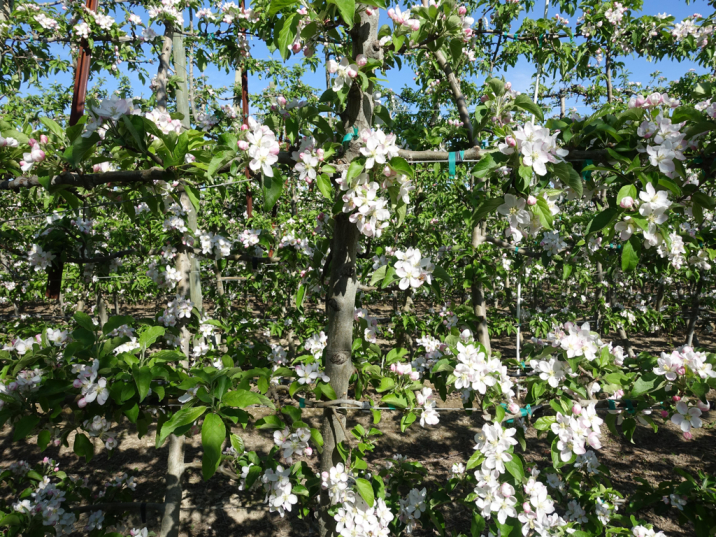}
    \includegraphics[width=0.4\textwidth]{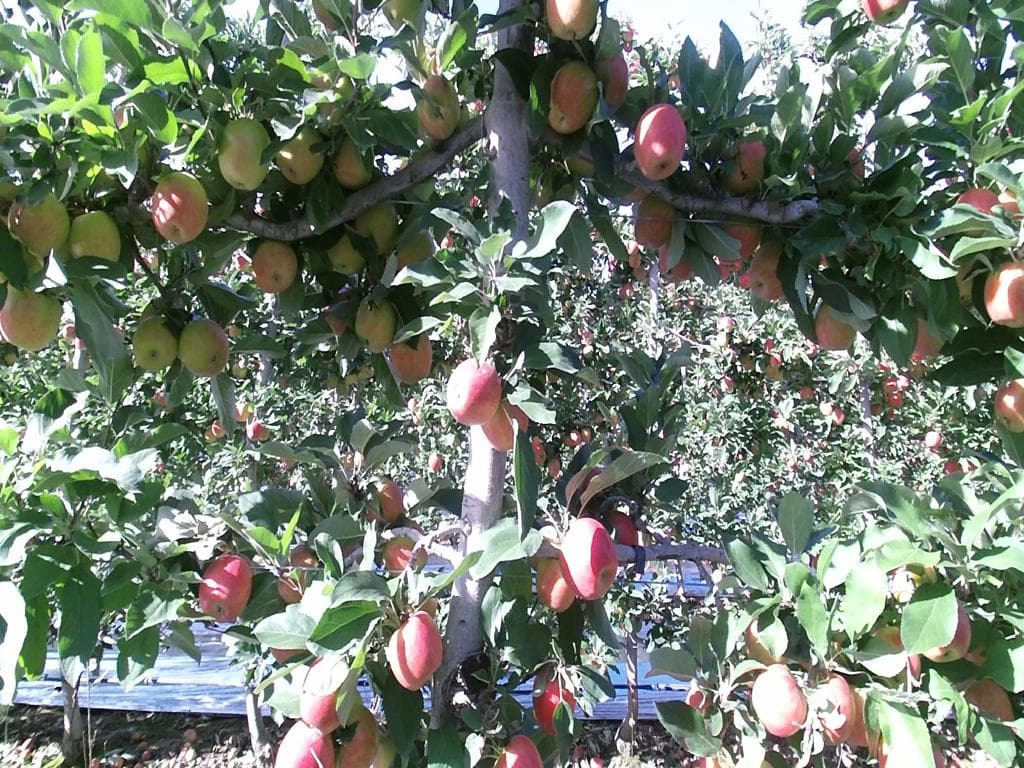}

    \caption{Example apple flower and fruit dataset used in this work. Apple flower dataset consisted of images acquired in two blooming stages (majority of flowers unopened and full bloom) in two apple canopy training architectures (Vertical wall, V trellis). Fruit images publicly available from \cite{gao2020multi} were annotated and evaluated.}
    \label{fig:Chap5_ExampleDataset}
\end{figure}

Robust computer vision algorithms, including deep learning, are increasingly being adopted in various agricultural applications, such as nutrient content assessment \cite{sulistyo2017deepintelligence_ase,zermas2020nitrogen_ase}, plant disease diagnosis \cite{cap2020leafgan_ase}, fruit detection, classiﬁcation,
and segmentation \cite{gao2020multi,wan2020faster,tian2019apple,gene2019multi}, ﬂower detection and segmentation \cite{farjon2019detection,bhattarai2020automatic,tian2020instance} in
challenging ﬁeld environments.
The fully supervised deep learning approaches offer high accuracy, precise object detection, and boundary estimation, which is well suited for applications such as robotic harvesting and fruit growth tracking. However, the detection-based approaches are highly complicated and may not necessary for many agricultural tasks such as crop-load estimation and bloom density estimation. For instance, full boom flowers in both stone (e.g., cherry) and pome fruits (e.g., apples) are densely located in constrained space with high flower-to-flower occlusions making individual flower annotation and detection challenging (see Fig. \ref{fig:Chap5_ExampleDataset}). Additionally, the detection of unopened flowers (see Fig. \ref{fig:Chap5_ExampleDataset}, top-left) is highly challenging since flowers are smaller in size and lack feature information.

In this work, we investigated the implementation of computer vision and deep learning in dense, highly occluded flower and fruit density estimation, localization, and counting, which is an under-explored area in agriculture. Density estimation has been studied in other application areas, specifically in crowd counting \cite{Li_2018_csrnet,wang2019sfcn,gao2019scar}. We argue a similar approach could be translated and extended into tree fruit production applications and help growers with efficient fruit production and management. This article proposed a segmentation-assisted deep regression network for flower and fruit density estimation, count, and localization leveraging a point annotation technique. The proposed approach does not require precise object boundary annotation, and bypasses object detection, region proposal estimation, and non-maximal suppression, making the estimation process simpler and computationally lighter. Integration of point annotation simplified the manual annotation process requiring only a pixel for each object, making the proposed method applicable even for highly occluded environments. The main contributions of the article are as follows: 
\begin{itemize}
    \item Inspired by U-Net \cite{ronneberger2015u} architecture and the recent advancement of ConvNeXt - T \cite{liu2022convnext}, a regression-based U-shaped network was developed by modifying ConvNeXt - T architecture for flower and fruit density estimation and counting, followed by a post-processing technique for localization. 
    \item A segmentation pipeline and attention mechanism were incorporated to handle unique agricultural problems caused by occlusions from leaves, branches, trunks, and the background objects (flowers and fruits in the immediate rows), which often appear similar to the objects of interest in terms of shape, size, and texture. The segmentation pipeline and attention mechanism highlighted and integrated the crucial flower and fruit features  while suppressing the non-relevant background information.
    \item The proposed approach was evaluated in challenging datasets acquired in unstructured commercial apple orchards during flower thinning and harvesting seasons, making the algorithm directly applicable to the field operations in different growing stages.
\end{itemize} 

The remaining parts of the paper are organized as follows. Section II presents detection-based approaches that can be extended for flower and fruit density estimation, counting, localization, and counting without detection methods. The technical details of the proposed approach are discussed in section III. Section IV presents the implementation and experimental details. Section V presents the results and discussion. Finally, we derive a few conclusions and suggest future directions in section VI. 

\section{Previous Work}
The classical learning approaches requiring extensive feature engineering could only support input data transformation into very few successive representation spaces  and showed poor performance in complex problems, which eventually gave rise to deep learning \cite {krizhevsky2017imagenet}. Dias et al. \cite{dias2018apple,dias2018multispecies} reported the implementation of deep learning in flower cluster segmentation using Convolution Neural Network (CNN) for feature extraction and region refinement approaches.  Bargoti and Underwood \cite{bargoti2017image} compared multilayer perceptron and CNN for image segmentation, followed by a watershed algorithm and circular Hough transform (CHT) for fruit detection. Additionally, popular end-to-end object detection and segmentation approaches such as Faster R-CNN (FRCNN) \cite {ren2015faster}, Mask R-CNN (MRCNN) \cite {he2017mask}, and You Only Look Once (YOLO) \cite {redmon2016you} have been heavily adopted in object detection and segmentation in agricultural contexts. FRCNN has been used for single to multi-class classification and detection of fruits such as apples \cite{gao2020multi}, kiwis \cite{wan2020faster}, oranges \cite{wan2020faster}, mangoes \cite{wan2020faster}, and flower clusters\cite{farjon2019detection}.  In another work, Farjon et al. \cite{farjon2019detection} proposed FRCNN-based cluster detection \cite{farjon2019detection} and Wang et al. \cite{wang2020side} proposed cluster segmentation-based methods for flower density estimation to optimize chemical flower thinning. Tian et al. \cite{tian2019apple} implemented improved YOLO v3 with DenseNet as a feature extractor for apple fruit detection. MRCNN, capable of simultaneous detection and instance-level segmentation, has been adopted for apple fruit, flower, and flower cluster detection and segmentation \cite{gene2020fruit,tian2020instance,bhattarai2020automatic}. Additionally, some previously reported approaches perform fruit counting using two-stage computation \cite{chen2017counting,hani2018apple,hani2020comparative}. The possible fruit locations were first separated by using a blob detection convolution network \cite{chen2017counting}, or region proposal network \cite{hani2018apple,hani2020comparative} followed by linear regression \cite{chen2017counting}, or classification method for counting \cite{hani2018apple,hani2020comparative}. 

Detection-based approaches are robust and accurate since the deep neural networks are trained on exact object features specified by polygons or bounding boxes. However, the detection-based approaches may not be suitable if the number of  object instances is large, the object size is small, or if a heavy occlusion exists between the object of interest and the object and background.  A study reported by Gomez et al. \cite{gomez2021deep} showed that detection-based and regression-based counting showed comparative results for low-density objects in agriculture datasets. However, for high-density objects (more than 100 objects/image), detection-based counting had shown five times larger error compared to regression-based counting \cite{gomez2021deep}. Furthermore, during full bloom, individual flower detection becomes highly challenging since flowers are located in dense clusters with substantial flower-to-flower occlusions. The object detection approaches have shown to perform poorly in detecting small objects \cite{huang2017speed}. Despite their maturity (unopened/fully opened), individual flowers have a substantially lower pixel footprint compared to well-sized apples. Furthermore, flowers are more randomly oriented than fruits, resulting in random  cluster structures and orientation, adding challenges even for human annotators (see Fig. \ref{fig:Chap5_ExampleDataset}) to annotate individual flowers. Hence developing computer vision approaches with the following characteristics is important: i) Ability to operate effectively on densely located objects without being influenced by the number of objects present in the image. ii) Simplified annotation process to facilitate easier and more efficient data labeling and network training. iii) Lightweight design enabling implementation on low-computation devices like cell phones and IoT devices.

Despite significant research in flower and fruit detection, limited studies have been reported for direct flower and fruit density estimation (sometimes referred to as mapping), localization, and counting. Recently, Farjon et al. \cite{farjon2021leaf} proposed a leaf-counting regression network using ResNet-50 and a feature pyramid network for banana and tobacco leaf counting and localization. Rahnemoonfar and Sheppard \cite{rahnemoonfar2017deep} used a modified version of Inception-ResNet trained in synthetic images to count red tomatoes. Our previous work studied direct flower and fruit counting and investigated the characteristic features contributing to the object count \cite{bhattarai2022weakly}. The proposed approach in this work is lightweight convolution neural network and leverages the ConvNeX - T architecture, an improvement over the ResNet-50 architecture. Furthermore, the proposed approach employs a U-Net-like structure with a skip connection allowing the low-level features to be transmitted to the deeper convolution layers. Furthermore, unlike the previously reported approaches, we introduce a segmentation branch and utilize the attention mechanism to suppress the background features (caused by a canopy in the immediate back row) and handle occlusions (caused by trunk, branch, leaves, and trellis wire) often seen in agriculture which is expected to improve the system performance while minimally adding computational overhead.

\section{Proposed Approach}
The proposed approach was to estimate object density maps specifying the flower and fruit distribution followed by object count and location estimation. To generate the density map, individual flowers and fruits were first annotated\footnote{http://www.robots.ox.ac.uk/~vgg/software/via/} as single point/pixel specifying the flower and fruit centroid. Only the flowers and fruits on the immediate front canopy row were annotated, and the flowers/fruits on the back canopy were considered as background. Since the single point/pixel provided minimal object feature information, two ground truth maps, \textit{1) Density Map} and \textit{2) Segmentation Map}, were generated by inclusing pixels in neighborhood of the annotated centroids to train the neural network. The density map was the final output of the proposed network, while the segmentation map acted as an intermediate output and was used to refine the density map.

\subsection{Ground Truth Generation}
\subsubsection{Density Map Generation}
Following the strategy used by \cite{lempitsky2010learning}, density maps were created by convolving normalized Gaussian kernel at each centroid. Since the area under the curve of the normalized Gaussian kernel equals one, the summation of all the pixels in the convolved image provided the total number of flowers and fruits. The distribution/spread of the Gaussian kernel could be varied by varying the standard deviation. Since the fixed standard deviation created density maps without regard to the distribution of neighboring objects, the adaptive standard deviation was computed using the KNN approach. The neighborhood search was performed in k-d tree space, and the standard deviation was defined as 15\% (flower) and 25\% (fruit) of  the distance from the nearest neighbor. Let $\delta(x-x_f^m,y-y_f^n); m \in \textbf{H},n \in \textbf{W},f \in I$ represent $f _{th} $ flower or fruit centroids located at pixel $(x^m,y^n)$. The 2D object density map was computed as
\begin{align}
D(\boldsymbol{x},\boldsymbol{y})=\sum_{f=1}^I \frac{1}{\sqrt{2 \pi} \sigma_f}            \exp{\left (- \frac{(x-x_f^m)^2 + (y-y_f^n)^2}{2 \sigma_f^2}\right )}
\end{align}

$I$ is the total number of centroids in the image with height, $\textbf{H}$ and width $\textbf{W}$, and $\sigma_f$ is the standard deviation of the Gaussian filter for centroid index $f$.

\subsubsection{Segmentation Map Generation} The purpose of a segmentation map was to act as a gate to pass the relevant features while suppressing irrelevant features (e.g. apples and flowers in the background row, the background sky) in the density map via binary classification of pixels. Although it was named as segmentation map, the objective was not the target to delineate individual objects but to refine generated density map to highlight feature information of relevant objects. The segmentation map was generated by convolving circular disk structuring elements in each flower and fruit centroid. A circular object with a center at origin and radius $r$ can be represented in standard form as $x^2+y^2=r^2$. For the flower and fruit centroids distributed at different locations within the image, the 2D object segmentation map was computed as
\begin{align}
    S(\boldsymbol{x},\boldsymbol{y})=\sum_{f=1}^I (x-x_f^m)^2+(y-y_f^m)^2\leq r^2; r=2\sigma_f
\end{align}

\begin{figure*}
\centering
\includegraphics[scale=0.8]{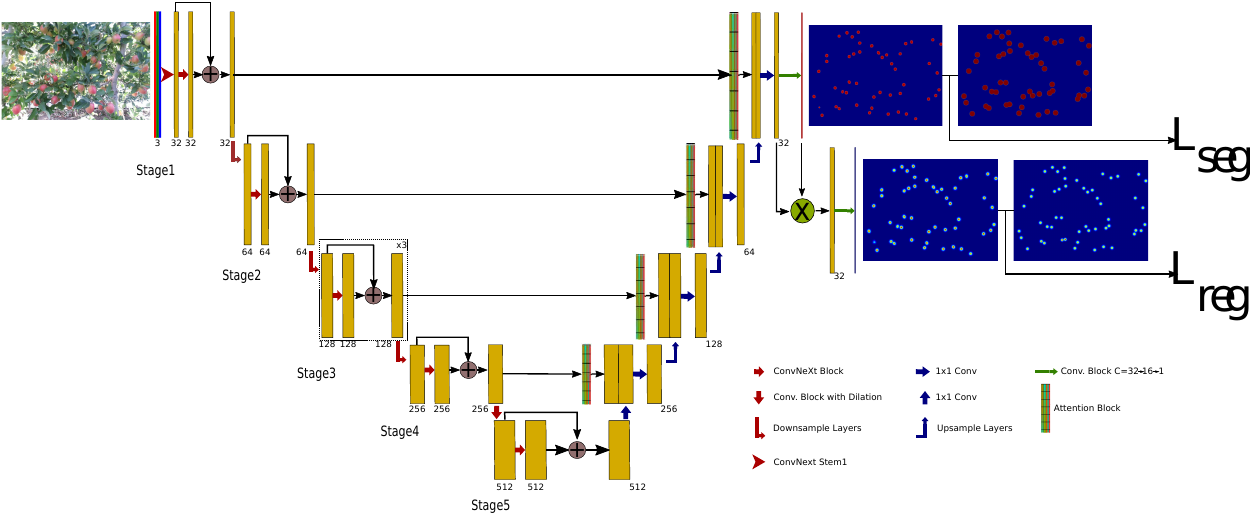}

\caption{AgRegNet: The proposed approach for flower and fruit density estimation in agriculture dataset. The network has U-Net inspired encoder-decoder framework with modified ConvNeXt - T in the encoder. A segmentation branch was added to transmit relevant features and suppress background information}
\label{Chap5_proposed_approach}
\end{figure*}
\subsection{Network Architecture}
The architecture of the proposed density estimation network was inspired by the encoder-decoder framework of the original U-Net architecture used for medical image segmentation \cite{ronneberger2015u}. Fig. \ref{Chap5_proposed_approach} illustrates the proposed AgRegNet architecture. The front-end feature extraction module of AgRegNet was designed by modifying the ConvNeXt-T network architecture originally derived from the ResNet-50 \cite{liu2022convnext}.  A dual attention mechanism proposed by \cite{woo2018cbam} was added within the modified ConvNeXt-T encoder skip connections and the encoder to decoder skip connections to enhance relevant features in both spatial and channel domains. Furthermore, a segmentation map was generated in the decoder section and was used as a feature filtering gate to pass the relevant features to the final density map estimation pipeline while suppressing irrelevant features.


\subsubsection{Front-end Feature Extraction Module}

The encoder section of the front-end feature extraction module was designed by modifying the ConvNeXt-T network architecture proposed by \cite{liu2022convnext}. During the preliminary experiment, the original ConvNext-T was implemented without any modifications. The original design of ConvNeXt-T aggressively reduced the feature map size to 1/32 of the original image size with heavy downsampling. The ConvNeXt-T architecture was modified so that the minimum feature map size on the encoder side after the final downsampling step would be 1/8 of the input image size. Additionally, deeper networks have a large number of hyperparameters and therefore require a large number of training images for improved generalization of the model. In agricultural situation where smaller dataset size is common, these networks are prone to overfitting \cite{gomez2021deep,bejani2021systematic}. Hence, the stage computation of the original ConvNext-T was modified from $3:3:9:3$ and the feature map dimensions of $(96,192,384,768)$ to $1:1:3:1:1$ with the feature map dimension of $(32,64,128,256,512)$.  Since the residual blocks in AgRegNet in each stage were reduced by a factor of 3, an additional stage was added to the aggregate and enhance the features in \textit{stage4} without downsampling. While processing the feature maps from \textit{stage4} to \textit{stage5}, dilated convolutions were used, which increases the receptive field, and has shown improved performance in semantic segmentation tasks while preserving the input feature size \cite{yu2015multidilated}. For the given three-channel input RGB images with $1024 (W) \times 768 (H)$ pixel resolutions, the output feature map resolution after \textit{stage5} would be $128(W/8) \times 96(H/8)$ with 512 channels (see Fig. \ref{Chap5_proposed_approach}). Apart from the modification in stage computation ratio and corresponding feature map dimensions, other design choices were kept similar to the original ConvNeXt-T \cite{liu2022convnext}. With the proposed modification, the size of the proposed AgRegNet was substantially reduced with a total of 9.45M trainable parameters compared to the original ConvNeXt-T with 29M parameters. 

\begin{figure}[!b]
\centering
\includegraphics[scale=0.42]{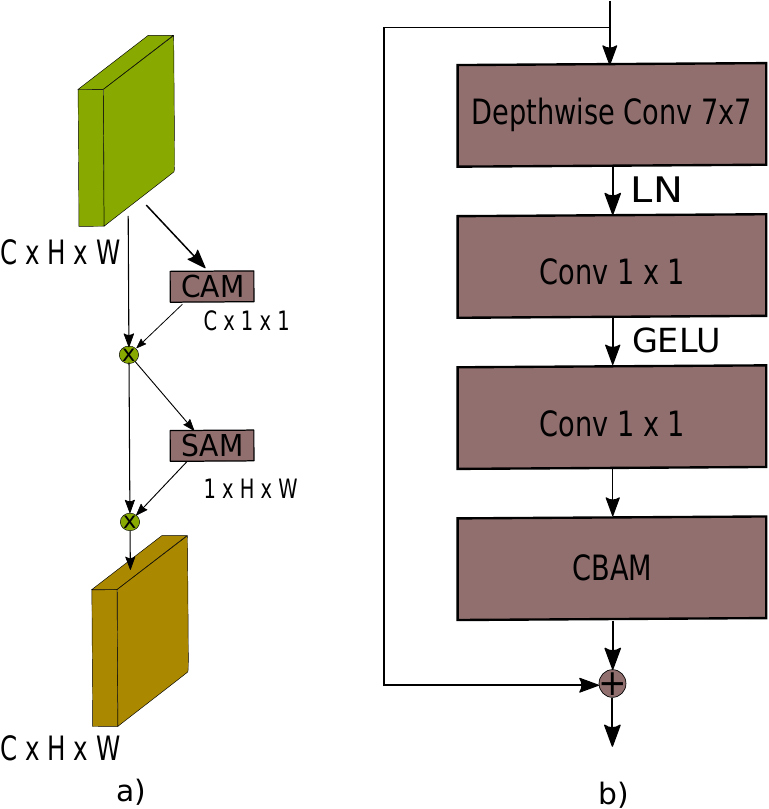}

\caption{a) Convolutional Block Attention Module (CBAM)\cite{woo2018cbam}. $\otimes$ represents elementwise multiplication; and b) Residual connection block in proposed modified ConvNeXt-T architecture }
\label{fig:Chap5_ResidualCBAMArchitecture}
\end{figure}
\subsubsection{Skip Connections and Feature Enhancement} As the proposed network leveraged the ConvNeXt-T and U-Net design, the skip connections were presented in two phases. The first set of skip connections were within the encoder module residual connection (see Fig. \ref{fig:Chap5_ResidualCBAMArchitecture}b) while the second skip connection spanned from each stage output of the encoder to the decoder module. Skip connections were commonly used as they improve detection and segmentation performance by propagating the  low-level features to deeper layers  while reducing the \textit{degradation problem}. In this work, the feature map propagated via skip connection was further enhanced by introducing Convolutional Block Attention Module (CBAM) proposed by \cite{woo2018cbam} (see Fig. \ref{fig:Chap5_ResidualCBAMArchitecture}a). CBAM acted as a feature enhancement and suppression gate in both spatial and channel dimensions. In CBAM, the Channel Attention Module (CAM) and Spatial Attention Module (SAM) were arranged sequentially, and both leveraged average-pool and max-pool operations for initial feature extraction.  Let $\boldsymbol{X} \in \mathbb{R}^{C \times H \times W}$ represent the input feature map, the CAM and SAM for CBAM were defined as
\begingroup\makeatletter\def\f@size{8.5}\check@mathfonts
\def\maketag@@@#1{\hbox{\m@th\large\normalfont#1}}
\begin{align}
    \boldsymbol{M_{CAM}}=\sigma(MLP(AvgPool(\boldsymbol{X}))+MLP(MaxPool(\boldsymbol{X}))) \\
    \boldsymbol{M_{SAM}}=\sigma(f^{7 \times 7}(Concat(AvgPool(\boldsymbol{X}),MaxPool(\boldsymbol{X})))
\end{align}
\endgroup
Where, $\boldsymbol{M_{CAM}} \in \mathbb{R}^{C \times 1 \times 1}$, and $\boldsymbol{M_{SAM}} \in \mathbb{R}^{1 \times H \times W}$. $MLP$ is multi-layer perceptron with one hidden layer, $\sigma$ is the Sigmoid activation function, $f^{7 \times 7}$ is the convolution layer with filter size 7.


\subsubsection{Backend Module}
\begin{figure}[!hb]
\centering
\includegraphics[scale=0.5]{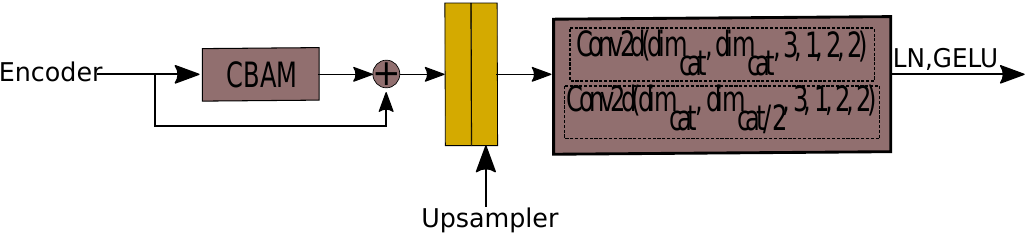}

\caption{Featuremap processing in each decoder block}
\label{fig:Chap5_DecoderProcess}
\end{figure}

The backend module worked as a post-processing operator to accumulate and highlight the feature information, and to compute the density map. Since the output feature map from the \textit{stage5} computation was equal to the size of the output feature map of \textit{stage4}, the feature maps were simply passed through a 1x1 convolution layer before concatenation. In the following decoder processing, the feature maps were combined via a concatenation, followed by a set of convolution operations (see Fig. \ref{fig:Chap5_DecoderProcess}). Since each stage of the encoder layer, except the last stage, was reduced to 1/2 of the size of the previous stage, feature maps were upsampled by factor of 2 before concatenating  in the decoder side. After concatenation, the feature information was further processed via convolution, normalization, and activation layers. Let, $Conv2d(C_i,C_o,k,s,p,d)$ represents 2d convolution with input channels $C_i$, output channels $C_o$, kernel size $k$, stride $s$, padding $p$, and dilation rate $d$. The decoder module after current layer concatenation and before immediate future layer concatenation is given by $[Conv2d(C_i=dim_{cat}, C_o=dim_{cat},k=3,p=d_{rate},d=d_{rate}), Conv2d(C_i=dim_{cat}, C_o=\frac{dim_{cat}}{2},k=3,p=d_{rate},d=d_{rate}),
LN,
GELU]$.
In this work, we set the dilation rate to 2.

\subsubsection{Segmentation and Density Map Generation}
The output of the decoder section in the backend module was further processed to generate two outputs that can be used to optimize the network using backpropagation. The segmentation map was similar to the semantic segmentation approach, which classified all the pixels into two classes: flower or fruit and the background. Given the input channels ($C_i$), the segmentation map was generated by 
$[Conv2d(C_i, C_i, k=3, p=d_{rate}, d=d_{rate}), Conv2d(C_i,C_o=1,k=3, p=d_{rate}, d=d_{rate}),
Sigmoid]$, where $d_{rate}=2$. The output segmentation map acted as the feature refinement gate for the density estimation network. The output feature map of the segmentation-based refinement pipeline was the elementwise multiplication of the segmentation map and the feature map to be refined. After refinement, the final density map was created by $[Conv2d(C_i, C_i, k=3, p=d_{rate}, d=d_{rate}), Conv2d(C_i,C_o=1,k=3, p=d_{rate}, d=d_{rate})]$, where $d_{rate}=2$.

\subsection{Count Estimation and Object Localization}
Object count was estimated by summing up the pixel intensity values of the density map. The predicted density map ($pred_{den}$) was post-processed to estimate the flower and fruit location. Algorithm \ref{Chap5_localizationalgorithm} was developed to estimate the flower and fruit centroid in the prediction map and associate the predicted  centroids ($pred_{peak}$) with the ground truth centroids ($gt_{peak}$). The density map estimated the likelihood that a pixel represented a flower or fruit. Hence, the higher the pixel intensity value, the greater the chances that a pixel is object centroid. For flower and fruit localization, the local peaks in the predicted density map were estimated using inbuilt $scikit-image$ package function $peak\_local\_max$. Once the local peaks or centroid of the predicted flower and fruit were estimated, a bipartite graph was created using the Euclidean distance between the ground truth and predicted centroids as a cost function. Finally, the Hungarian algorithm was used to perform one-to-one matching between the ground truth and predicted centroids minimizing the cost. 
\begin{algorithm}
\caption{Flower and fruit localization}
\begin{algorithmic}[1]
    \Require $pred_{den}$, $gt_{peak}$
    \Ensure{Association of $gt_{peak}$ and $pred_{peak}$ }
    \Statex
    \Function {EstimateGroundPredictLocPair}{}
        \State {Minimum distance separating peaks ($d_{min}$) $\gets$ 2}
        \State {$pred_{peak} \gets $ \textit{peak\_local\_max}$(pred_{den},d_{min})$}
        \State {$N \gets $ length ($gt_{peak}$), $M\gets$ length ($pred_{peak}$)}
        \State {Cost ($C$) $\gets$ 
                $\begin{bmatrix}
                    0.0 & \ldots & 0.0 \\
                    \vdots & \ddots & \vdots \\
                    0.0 & \ldots & 0.0 \\
                \end{bmatrix}_{N \times M}$}
        \For{$i \gets 1$ to $N$}
            \For{$j \gets 1$ to $M$}
                \State{$C [i][j]\gets$ Eucl.$(gt_{peak}[i],pred_{peak}[j])$}
            \EndFor
            \EndFor
        \State ($gt_{peak_{index}},pred_{peak_{index}})\gets$ \textit{bipartite\_match}$(C)$
        \State \Return $gt_{peak_{index}},pred_{peak_{index}}$
        \EndFunction
    
\end{algorithmic}
\label{Chap5_localizationalgorithm}
\end{algorithm}

\subsection{Loss Function}
The network loss was computed by combining the Dice loss for the segmentation map and Mean Squared Error (MSE) loss for the density map. In this work, the Dice coefficient \cite{milletari2016dice} was selected as a segmentation optimization metric since it has shown robust performance in handling class imbalance problems compared to Binary Cross Entropy loss. Class imbalance was common in the flower and fruit dataset datasets where objects of interest occupy a substantially lower number of pixels compared to the background. The Dice coefficient is given as 

\begin{align}
    D_{coeff}=\frac{2\sum_{i=1}^ N p_{{seg}_i}g_{{seg}_i}}{
    \sum_{i=1}^N p_{{seg}_i}^2+\sum_{i=1}^N g_{{seg}_i}^2
    }
\end{align}
where $i \in \textbf{N}; \textbf{N}=H \times W$, and $g_{seg}$ and $p_{seg}$ are the ground truth and predicted binary segmentation maps. The segmentation loss was computed as $L_{seg}=1-D_{coeff}$. The density map estimation was optimized using Mean Squared Error (MSE) loss given as 
\begin{align}
    L_{reg}=\frac{1}{N}\sum_{i=1}^N (g_{den}-p_{den})^2
\end{align}
The total loss ($L_{loss}$) was computed as $L_{loss}=L_{den}+\alpha L_{seg}$. Where $\alpha=0.01$ was the scaling parameter computed empirically.

\section{Experimental Evaluation}

\subsection{Dataset}
The experimental evaluation consisted of two datasets with apple flowers and apples (see Table \ref{tab:Chap5_Dataset Details}). Apple flower images were collected over two growing seasons (2018 and 2019) using different imaging sensors. Images were collected from three commercial apple orchards in Washington, USA, with multiple varieties grown in fruiting wall canopy architecture. Trees were trained and pruned to create narrow 2D structures using vertical or V-shaped trellis systems in fruiting wall architectures. The flower dataset included images of early to late phases of the flower blooming in three apple varieties: Scifresh (Vertical Fruiting Wall), Envi (V trellis), and HoneyCrisp (V trellis). Additionally, the apple fruit dataset publicly released by \cite{gao2020multi} as a part of a multi-class fruit classification experiment was annotated and used. The fruit dataset consisted of images of 800 harvest-ready apple canopies in vertical fruiting wall architecture \cite{gao2020multi}. Images with similar appearances were identified using Structural Similarity Index Measure (SSIM) greater than 95\%, 
 which were later removed by human verification resulting in 630 unique apple canopy images\cite{hore2010image}. For experimental evaluation, each dataset was divided into a training set (75\%) and a test set (25\%). The validation dataset was obtained by randomly cropping the test images.
\begin{table}[!hb]
    \centering
    \setlength{\tabcolsep}{1.5pt}
    \caption{Dataset used in this study; Images were acquired from multiple apple varieties in different architectures and imaging sensors.}
    \label{tab:Chap5_Dataset Details}
    \scalebox{0.9}{
    \begin{tabular}{lclccc}
        \hline \hline
         Dataset   & \#  & Sensors(Resolution) & \multicolumn{3}{c}{Statistics}  \\
         \cline{4-6}
                         &Images           &                & Objects & Min/Max & Avg                  \\
        \hline 
        Apple Flower & 325     & ZED2,Kinect V2 (1920 $\times$ 1080)  & 43031   & 4/434    & 132               \\
        &&                             Sony RX100 (5472 $\times$ 3648) &&&                                         \\
        \hline
        Apple Fruit \cite{gao2020multi} & 630     & Kinect V2 (1920 $\times$ 1080) & 34404    & 10/121    & 55           \\

        \hline \hline
    \end{tabular}}
\end{table}

\subsection{Model Training}
The first four stages of the modified ConvNeXt-T network were initialized using a pre-trained ConvNeXt-T model. The rest of the layers were randomly initialized using a Gaussian distribution with a mean of 0 and a standard deviation of 0.01. The network was trained with Adam optimizer using a learning rate of 0.0004 and a learning rate decay rate of 0.995 at each epoch. Network training was conducted for up to 200 epochs using the PyTorch library using two Nvidia Titan X 12 GB GPUs. For all datasets, the training and test sets were reshaped to $1024(W) \times 768(H)$, while the validation set was obtained by randomly cropping test set images to $768(W) \times 576(H)$. 

\subsection{Performance Metrics}
The density maps' quality was evaluated using SSIM and Peak Signal to Noise Ratio (PSNR) \cite{hore2010image}. SSIM metric computed the similarity between two images in terms of the closeness of mean luminance, contrast, and structure. For structure comparison, the correlation between ground truth
and the predicted density map was estimated by computing the covariance matrix \cite{hore2010image}.PSNR, on the other hand, took into account the mean square error between the ground truth and the predicted density maps at a pixel level. A lower MSE value corresponded to a lower numerical difference between the image pixels resulting in a larger PSNR value.  For count evaluation, Mean Absolute Error (MAE) and Root Mean Squared Error (RMSE) were used.
\begin{align}
    MAE=\frac{1}{N} \sum_{i=1}^N \lvert C_i^{pred}-C_i^{gt} \rvert \\
    RMSE=\sqrt{\frac{1}{N} \sum_{i=1}^N {\lvert C_i^{pred}-C_i^{gt} \rvert}^2}
\end{align}

Where $N$ was the total number of images in the test set. $C_i^{pred}$ was predicted count, and $C_i^{gt}$ was ground truth count. To qualitatively evaluate the localization results, precision and recall metrics were estimated. Given the matched ground truth and predicted centroids, a cutoff value ($T$) was used to determine whether the matched centroids were correctly associated. The following criteria were used to evaluate the predicted centroids.
\begin{align}
    pred_{peak}\in \begin{cases}
    True Positive, &\text{if}\ d(gt_{peak},pred_{peak}) \leq T \\
    False Positive, & \text{otherwise}
    \end{cases}
\end{align}
The Average Precision (AP) and Average Recall (AR) were computed by varying the cutoff threshold $T$. Value of $T$ ranged from $\sigma_f$ to $2.2\sigma_f$ with an increment of 1. $\sigma_f$ was the default standard deviation of the Gaussian kernel used to create the ground truth density map. Furthermore, $d(gt_{peak},pred_{peak})$ was the Euclidean distance between the ground truth and the predicted peak location.

\section{Results and Discussion}
\subsection{Density Map Evaluation}
\begin{figure*}[!ht]
    
    \centering
    \includegraphics[width=0.19\textwidth]{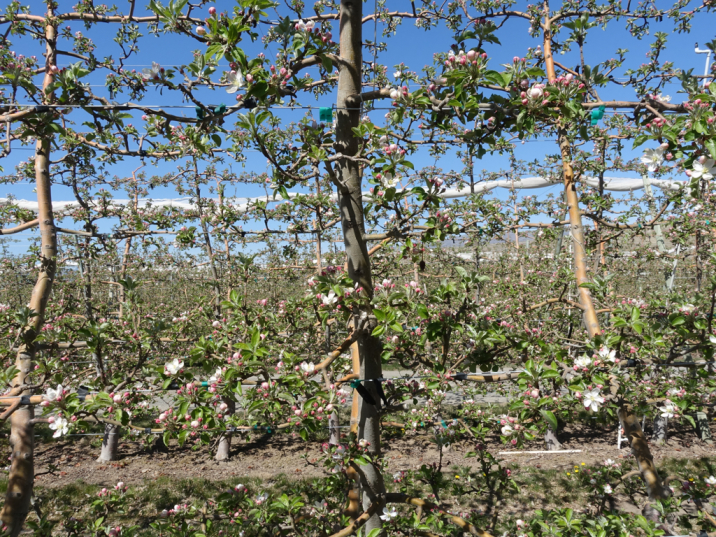}
    \includegraphics[width=0.19\textwidth]{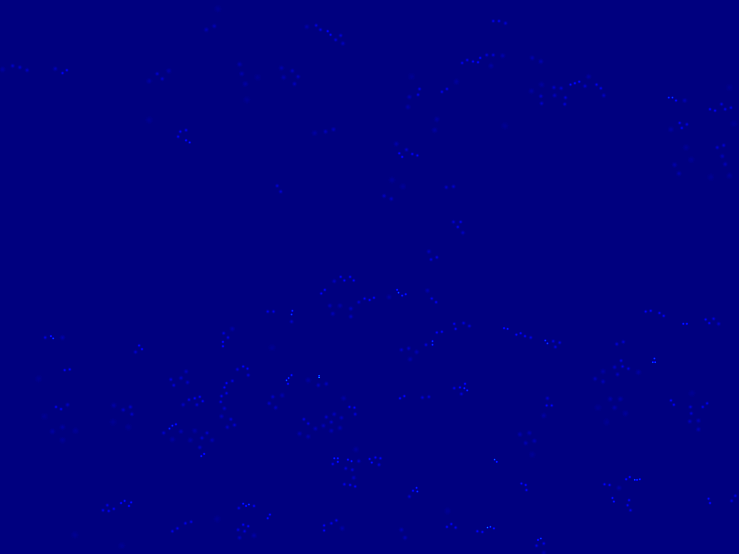}
    \includegraphics[width=0.19\textwidth]{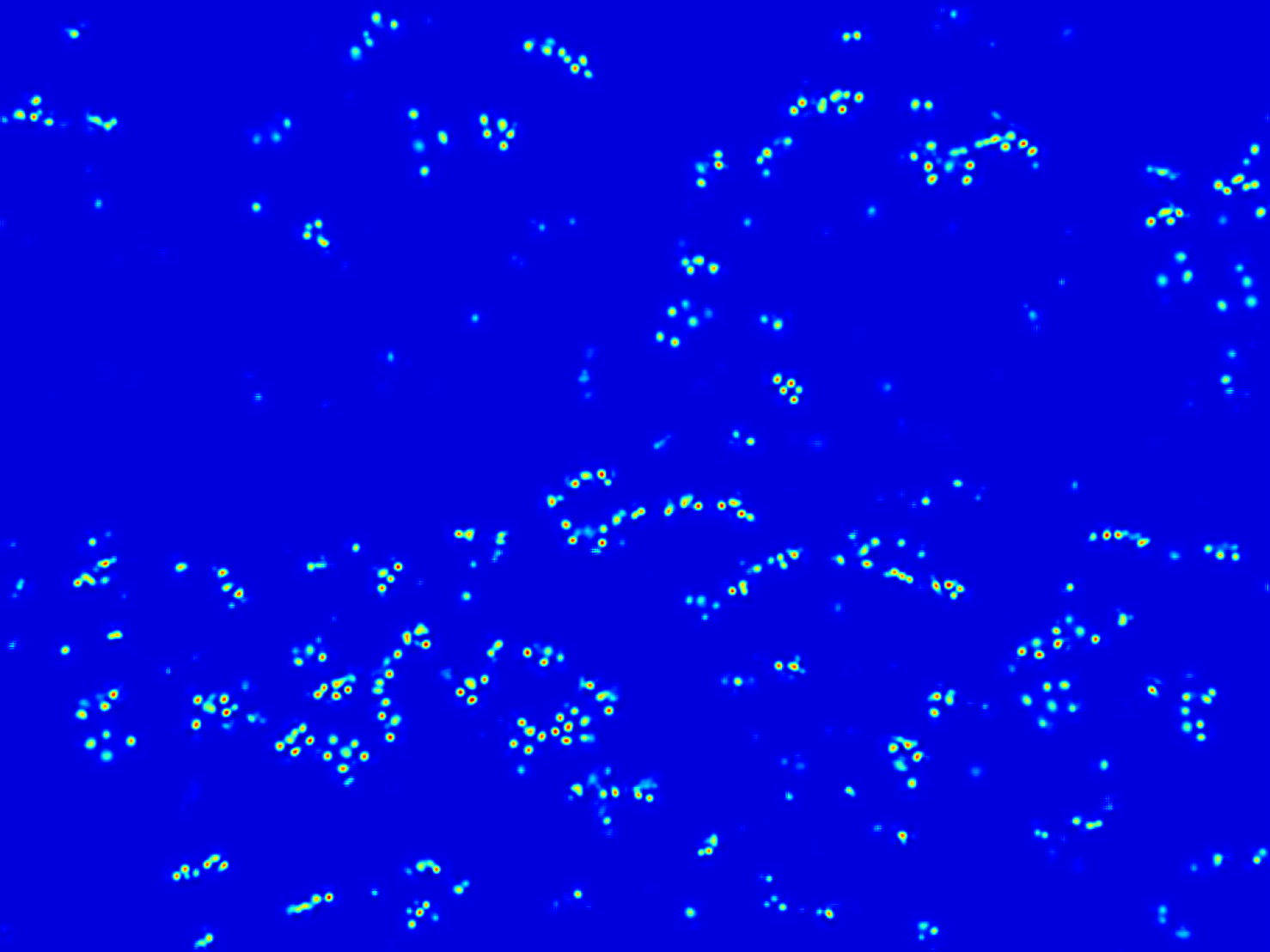}
    \includegraphics[width=0.19\textwidth]{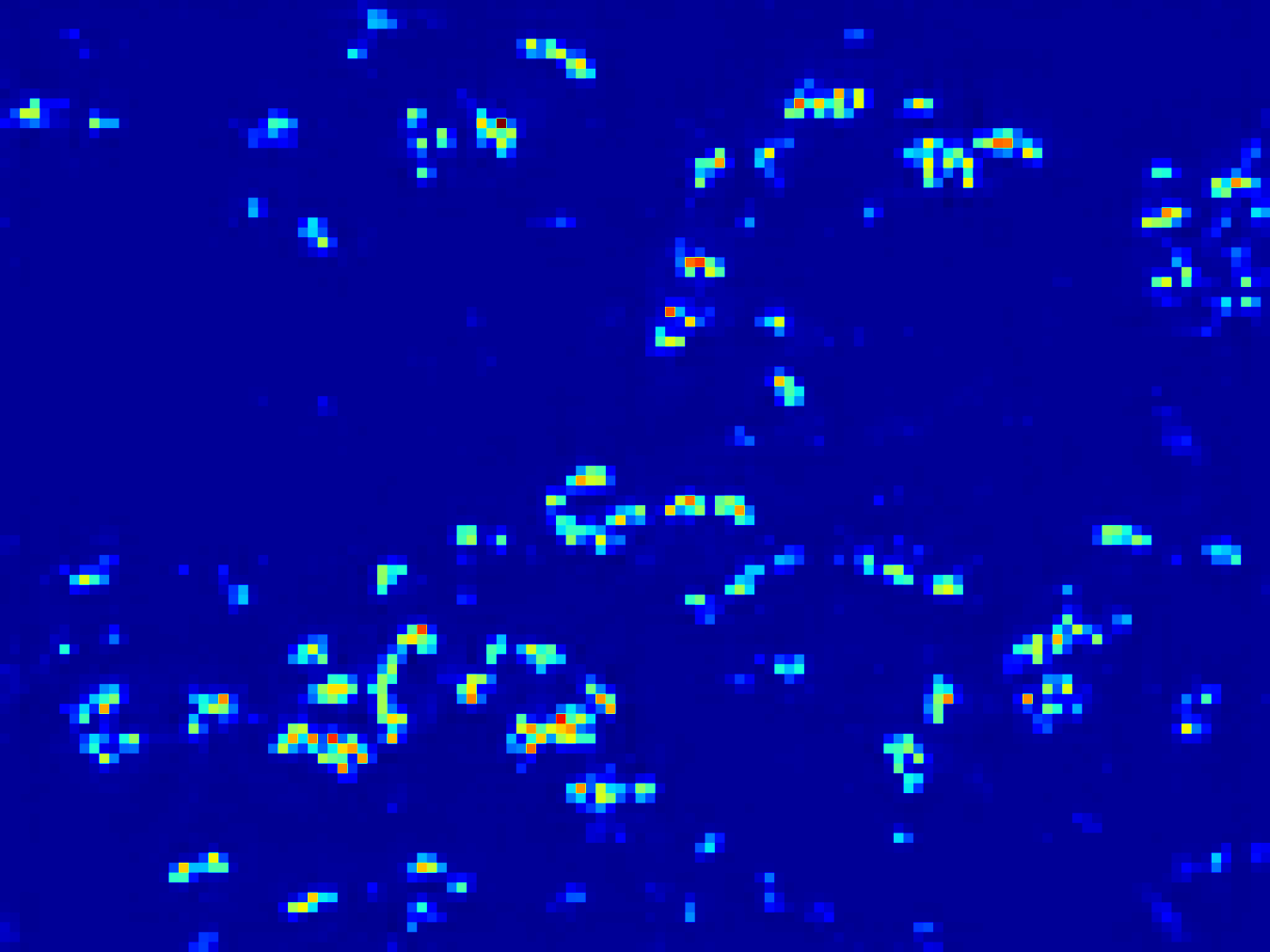}
    \includegraphics[width=0.19\textwidth]{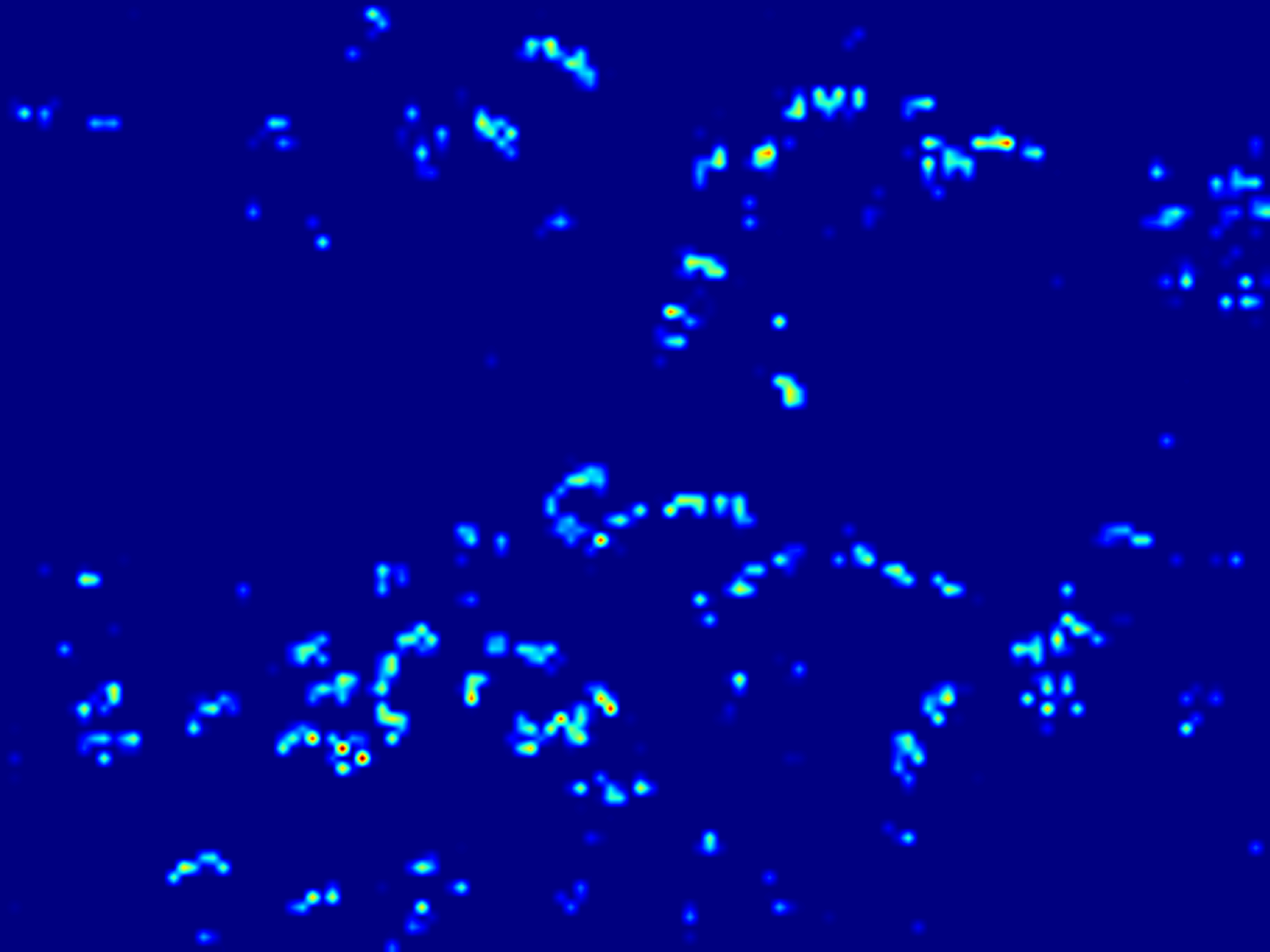}  
    
    \vspace{0.2mm}
    \centering
    \includegraphics[width=0.19\textwidth]{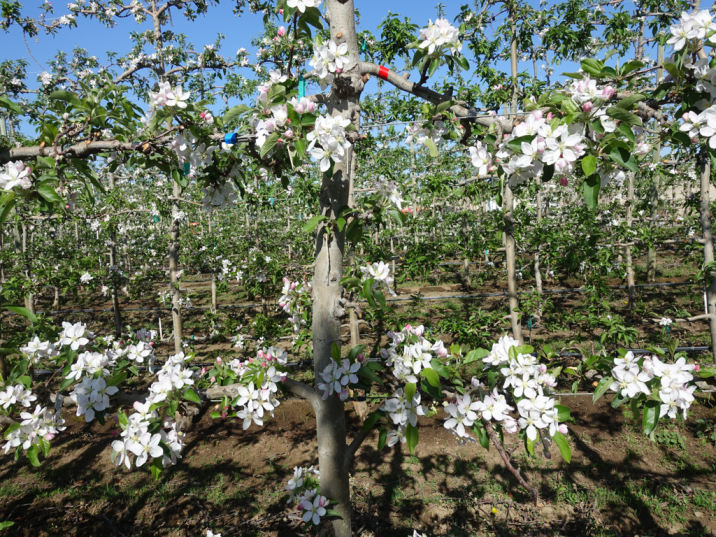}
    \includegraphics[width=0.19\textwidth]{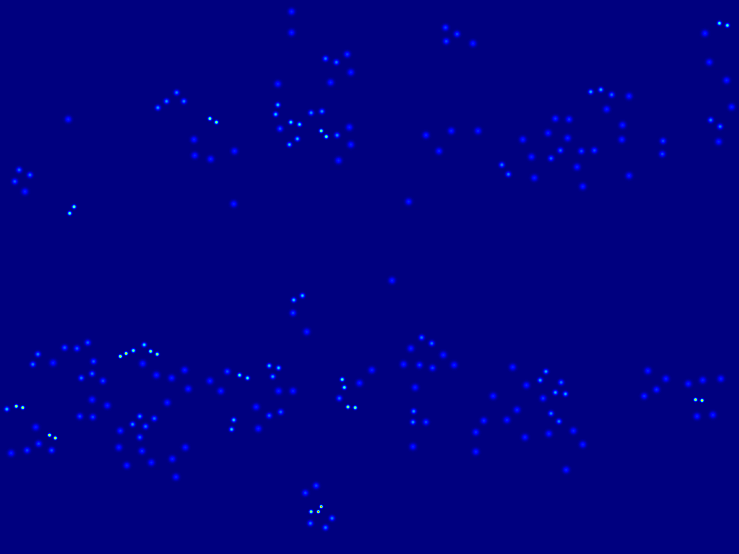}
    \includegraphics[width=0.19\textwidth]{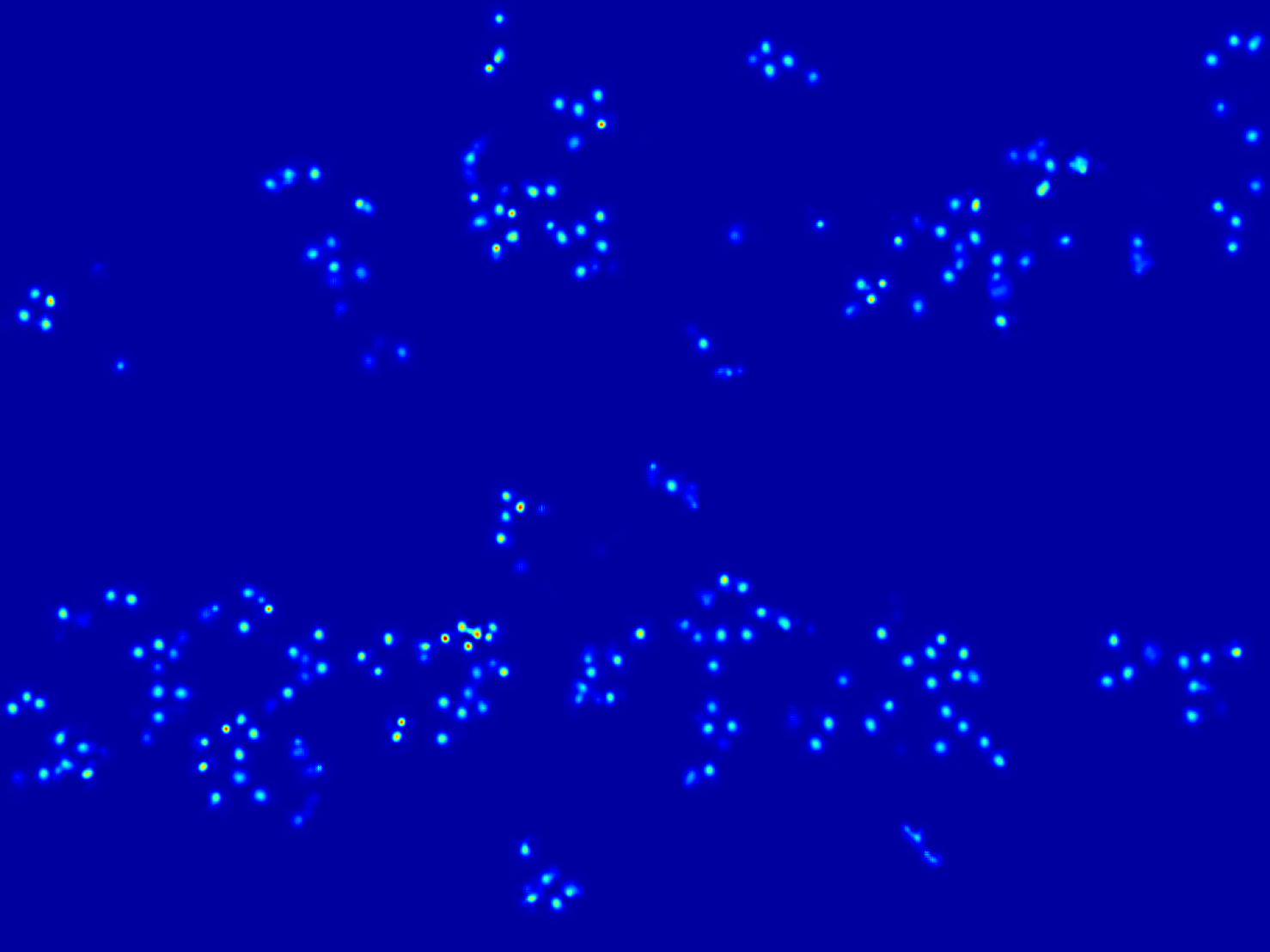}
    \includegraphics[width=0.19\textwidth]{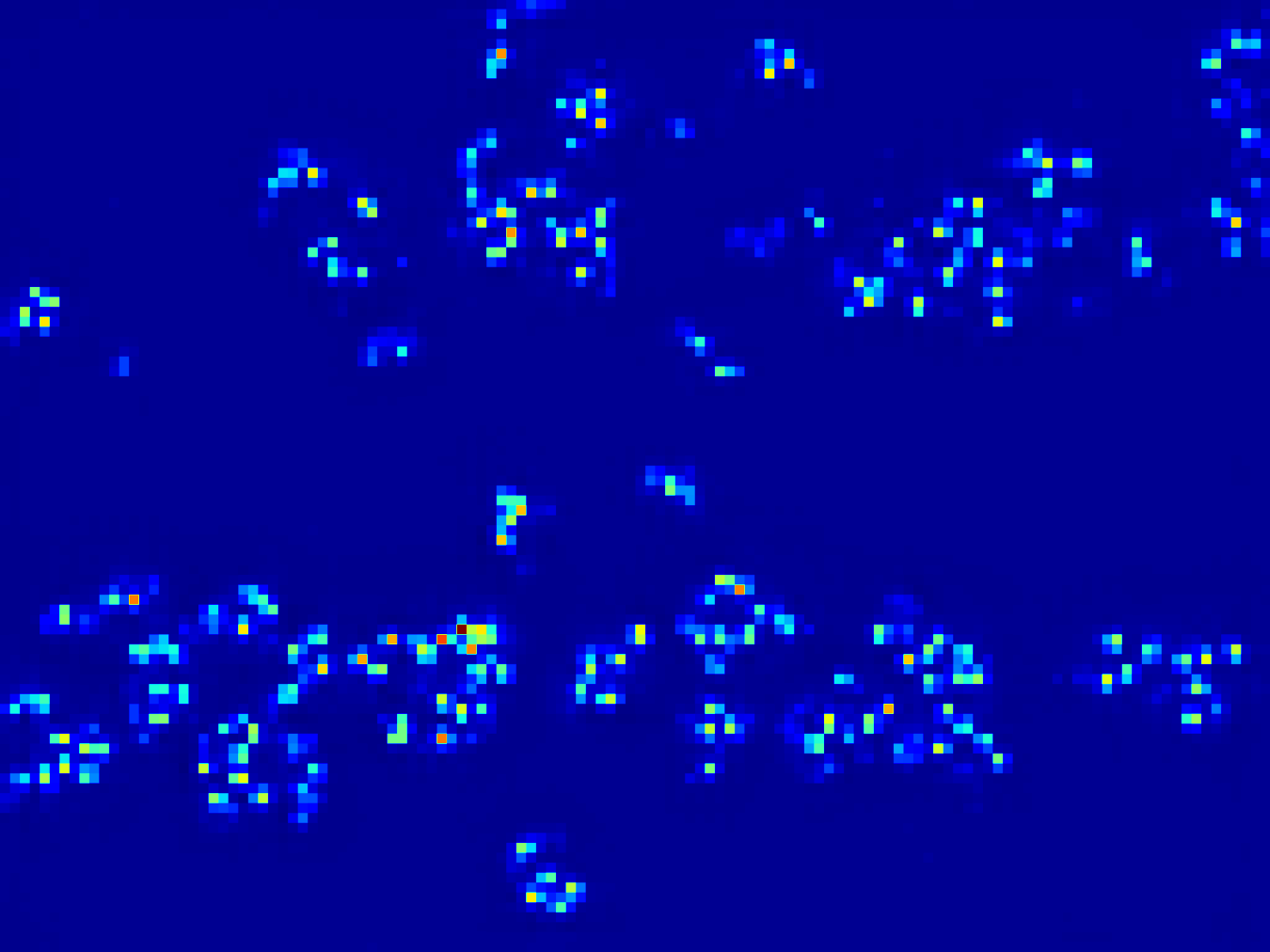}
    \includegraphics[width=0.19\textwidth]{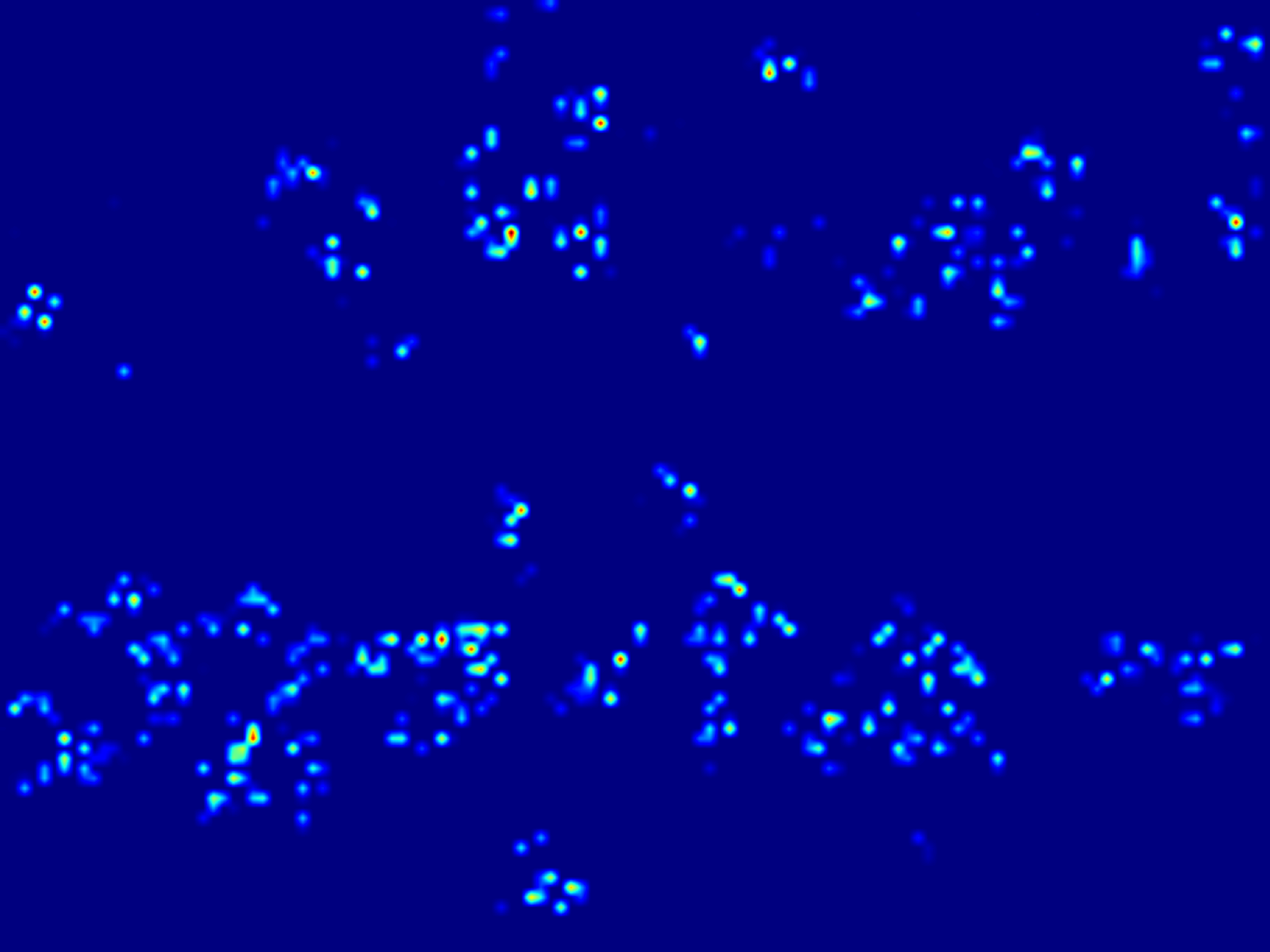}
    
    \vspace{0.2mm}
    \centering
    \includegraphics[width=0.19\textwidth]{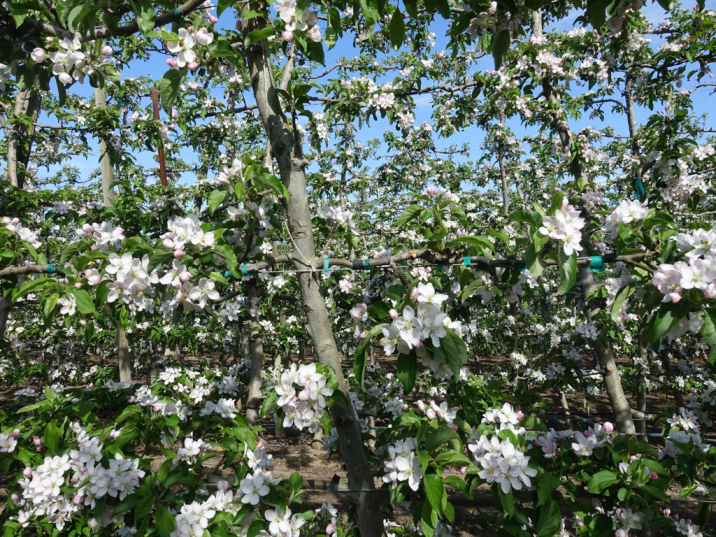}
    \includegraphics[width=0.19\textwidth]{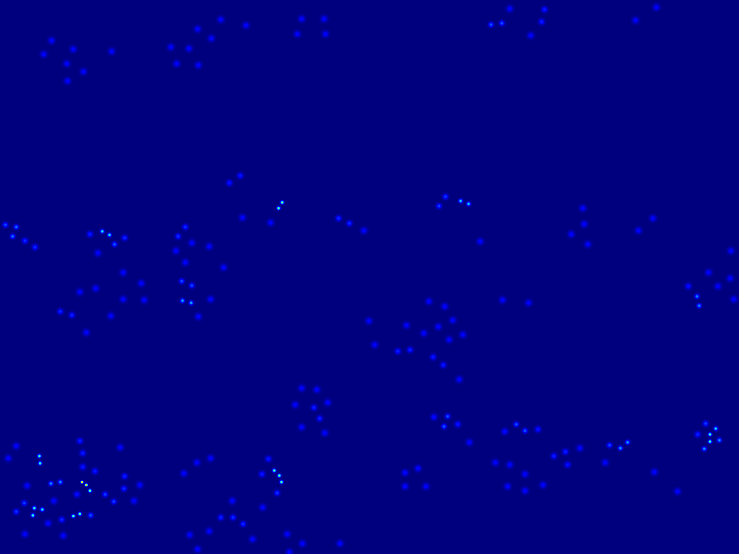}
    \includegraphics[width=0.19\textwidth]{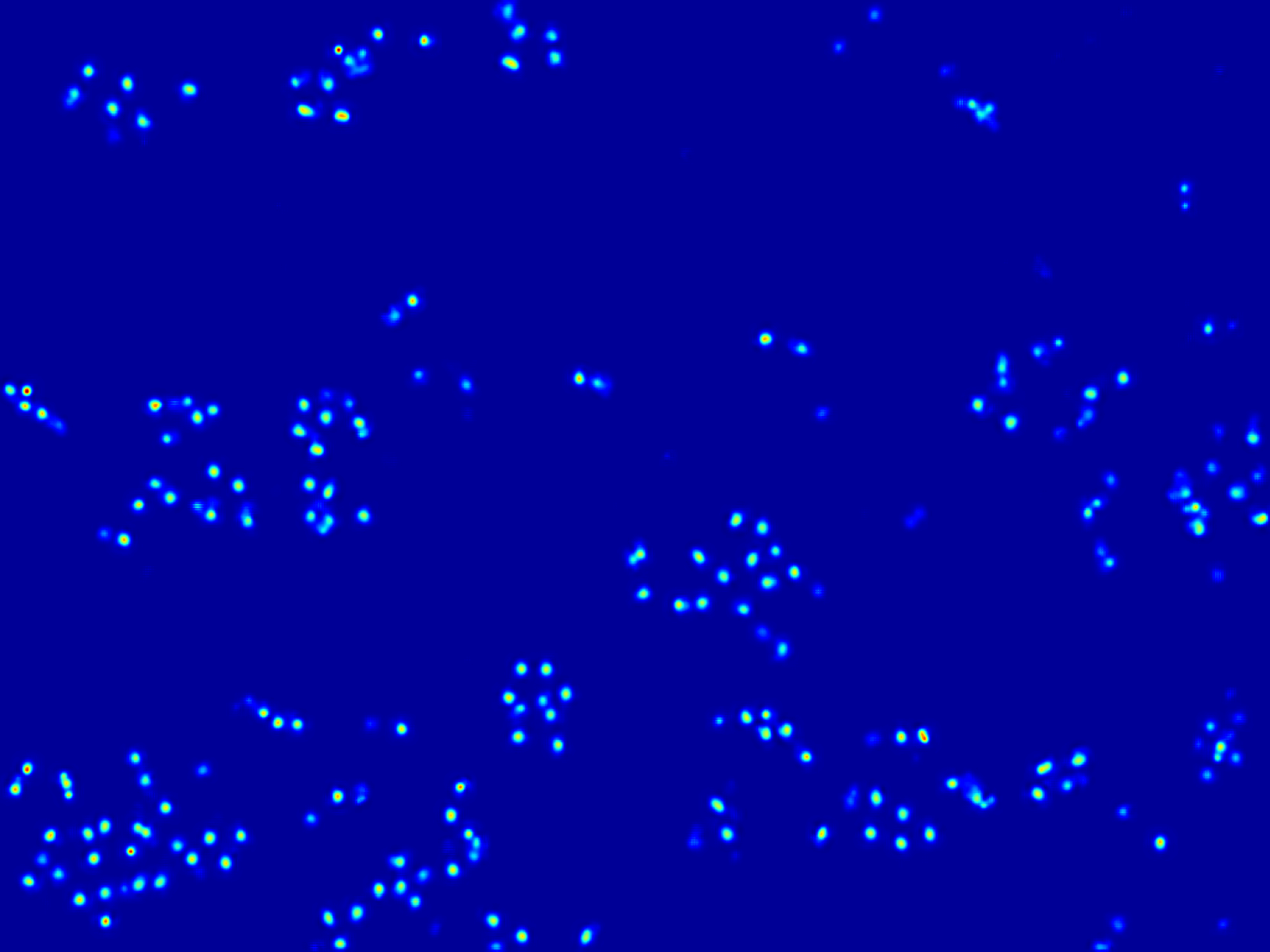}
    \includegraphics[width=0.19\textwidth]{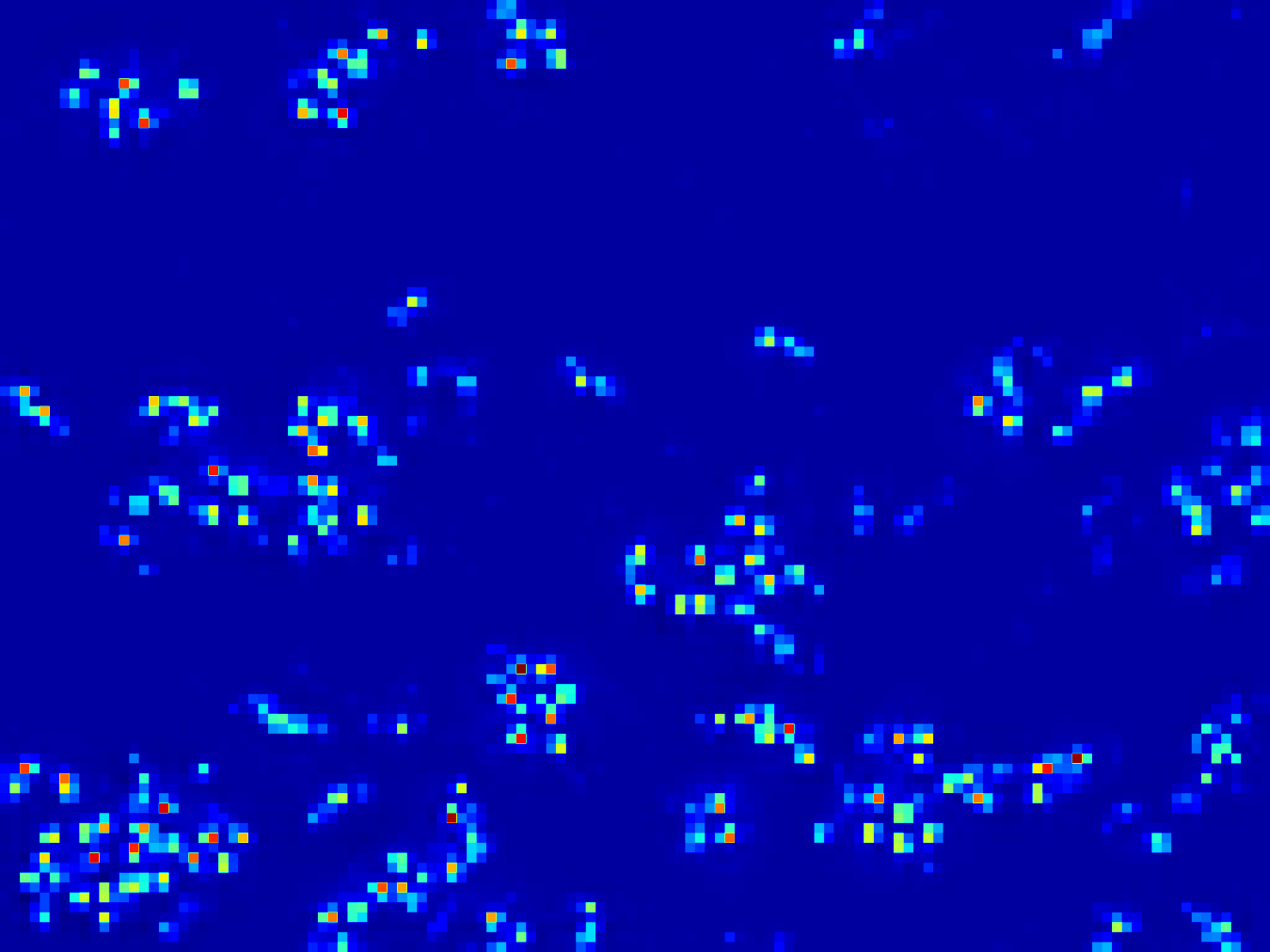}
    \includegraphics[width=0.19\textwidth]{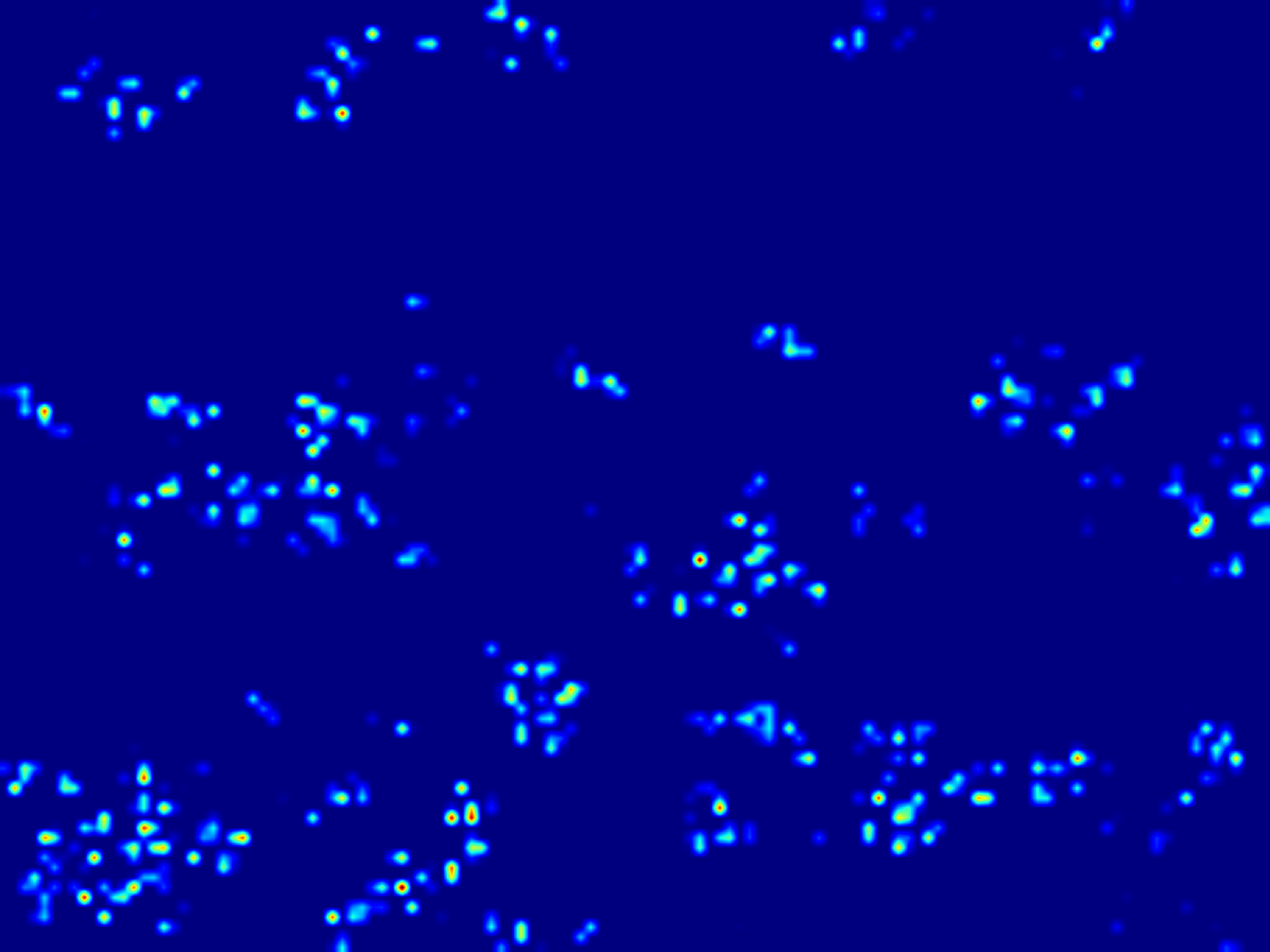}
    \vspace{0.02mm}
    \begin{minipage}{.19\linewidth}
    \centering
    \subfloat[Canopy images]{\label{main:a}\includegraphics[scale=0.085]{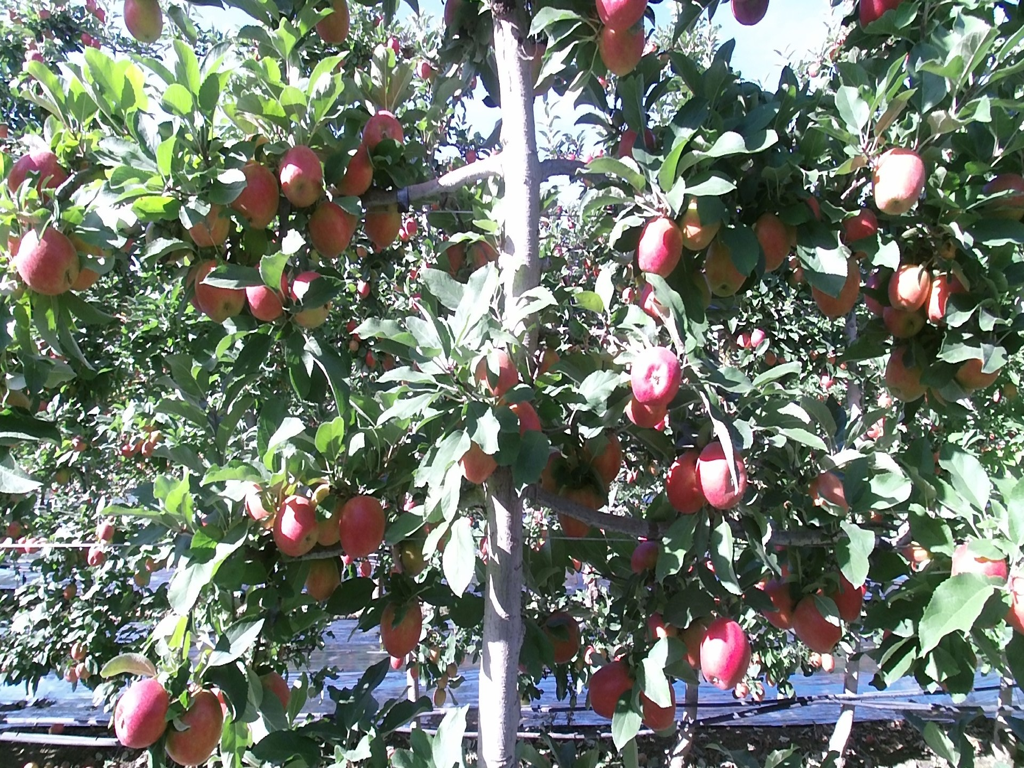}}
    \end{minipage}
    \begin{minipage}{.19\linewidth}
    \centering
    \subfloat[Ground truth]{\label{main:b}\includegraphics[scale=0.25]{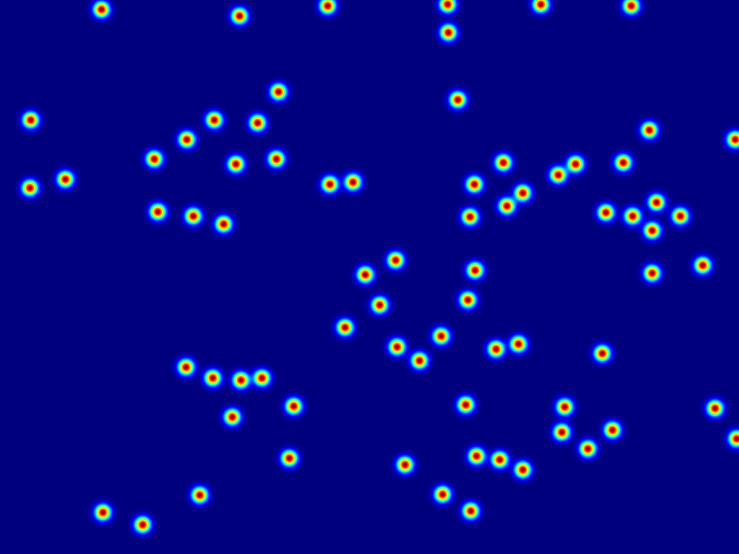}}
    \end{minipage}
    \begin{minipage}{.19\linewidth}
    \centering
    \subfloat[AgRegNet (Ours)]{\label{main:c}\includegraphics[scale=0.25]{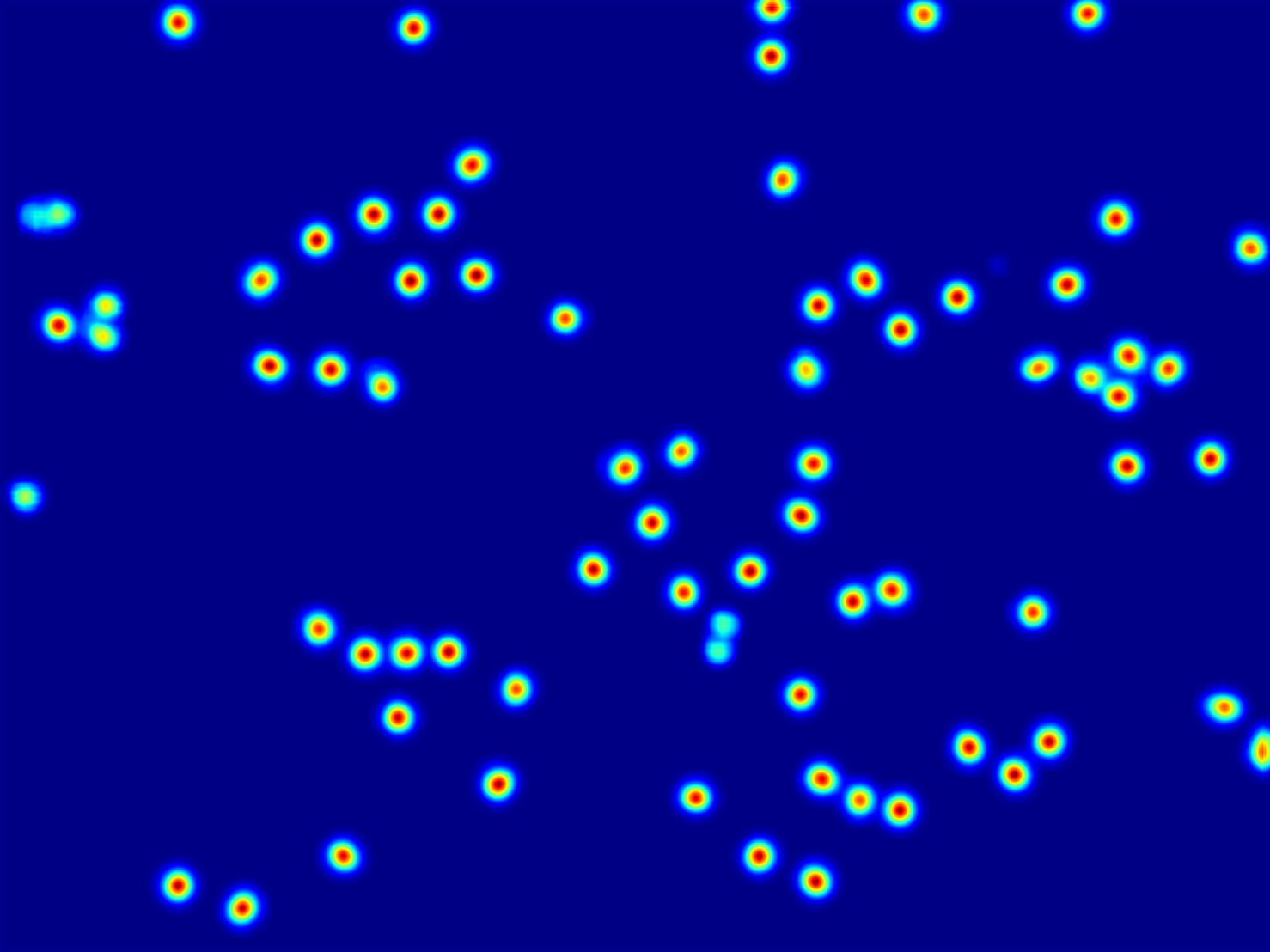}}
    \end{minipage}
    \begin{minipage}{.19\linewidth}
    \centering
    \subfloat[CSRNet]{\label{main:d}\includegraphics[scale=0.25]{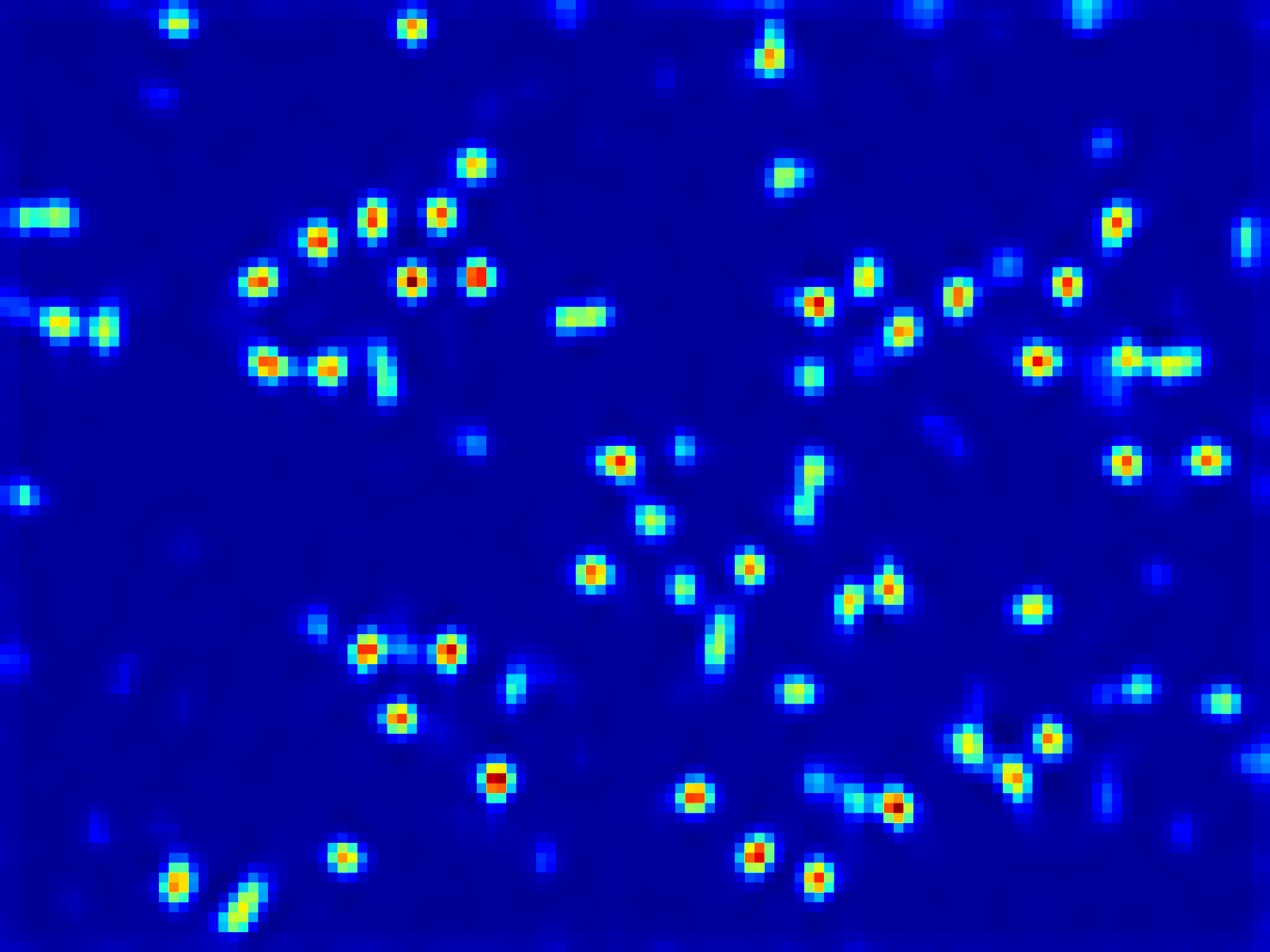}}
    \end{minipage}
    \begin{minipage}{.19\linewidth}
    \centering
    \subfloat[SCAR]{\label{main:a}\includegraphics[scale=.25]{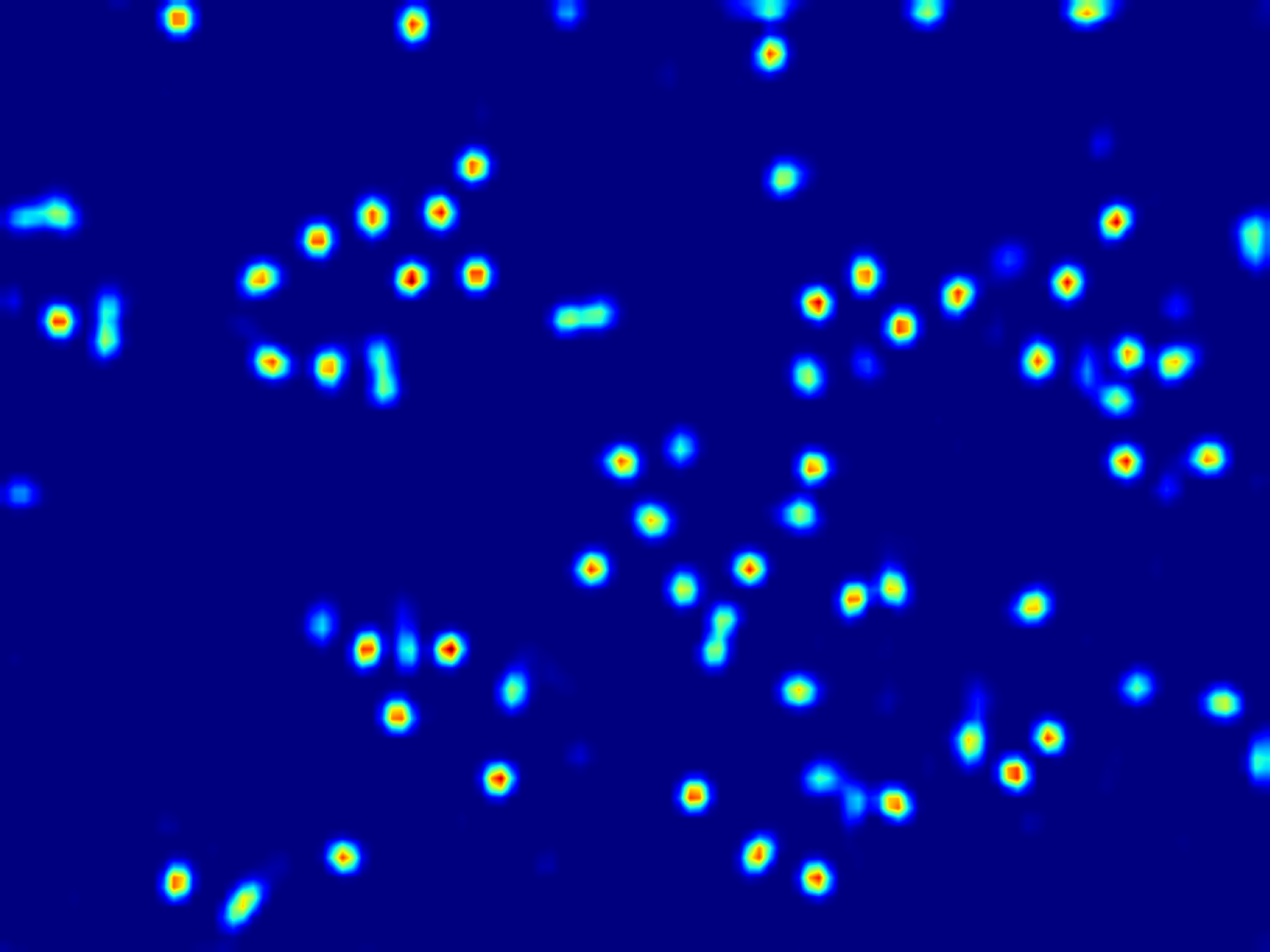}}
    \end{minipage}
    \caption{Qualitative comparison of the density maps of the proposed AgRegNet with other approaches}
    \label{fig:Chap5_allQualitative}
\end{figure*}
The proposed network was compared with different density map estimation approaches in terms of PSNR, and SSIM \cite{hore2010image}. Among the all evaluated approaches the proposed approach was lightweight with 9.45M total trainable parameters. The following were some of the popular approaches compared with the proposed approach.

\begin{itemize}
    \item CSRNet\footnote{https://github.com/leeyeehoo/CSRNet-pytorch} \cite{Li_2018_csrnet}: CSRNet, a popular population estimation method, used VGG-16 as the backbone and leveraged dilated convolution for performance improvement, followed by backend convolution block to compute density map.
    \item SFCN \footnote{https://github.com/gjy3035/GCC-SFCN}\cite{wang2019sfcn}: SFCN used a spatial fully convolution neural network stacked on the top of the ResNet101 backbone followed by a regression layer to compute the density map.
    \item SCAR\footnote{https://github.com/gjy3035/SCAR} \cite{gao2019scar}: SCAR used VGG16 backbone and Spatial and Channel Attention module similar to one used by  \cite{fu2019dual,goodfellowzhang2019sagan}
    
\end{itemize}
\begin{table}[!h]
    \centering
    \caption{Comparison of PSNR and SSIM of the proposed approach with different density map estimation methods.}
    \label{tab:Chap5_psnrSsimAllData}
    \begin{tabular}{lcccc}
        \hline  \hline
        Method &  \multicolumn{2}{l}{Apple Flower} & \multicolumn{2}{c}{Apple Fruit \cite{gao2020multi}} \\
        \cline{2-3} \cline{4-5} 
                        & PSNR & SSIM     & PSNR & SSIM  \\
        \hline 
        CSRNet\cite{Li_2018_csrnet}          &24.6 &0.845 &22.1 &0.734  \\
        SFCN \cite{wang2019sfcn} &28.5&0.912 &22.6&0.901 \\        
        SCAR \cite{gao2019scar} &27.8 &0.910 &\textbf{24.7} &0.908 \\        
        AgRegNet (Ours)          &\textbf{31.2} &\textbf{0.938}      &24.1 &\textbf{0.910} \\
        \hline \hline
    \end{tabular}
\end{table}

\begin{figure*}[!ht]
    
    \centering
    \includegraphics[width=0.85\textwidth]{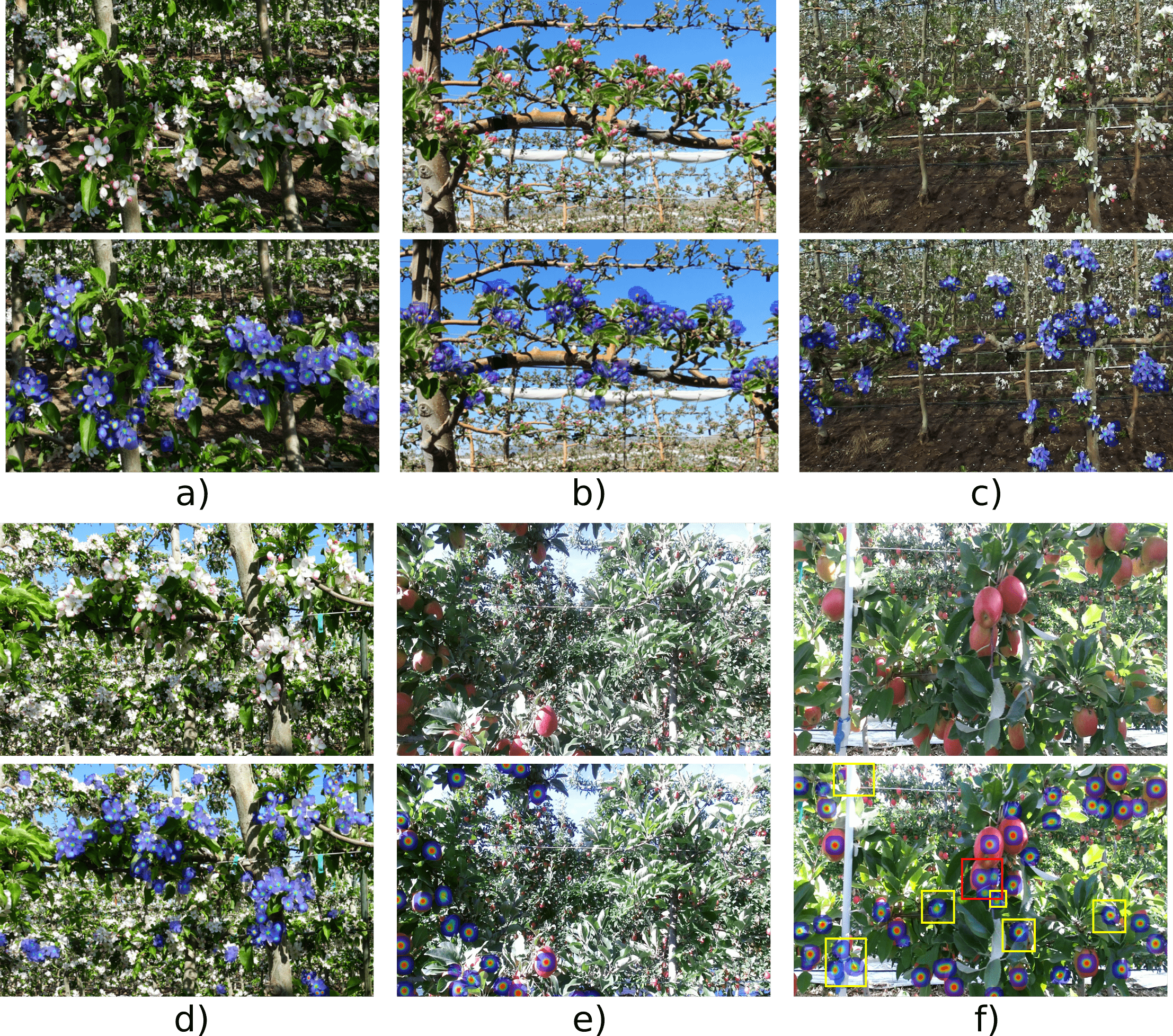}
    \caption{Different flower fruit density estimation scenarios; a) Most of the clusters were in full bloom, although there were some clusters that only have the \textit{King Flower} open; b) Early flowering stage with the majority of the flowers not yet open; c) Full bloom clusters and un-opened clusters in same canopy; d) V-trellised system with background flowers in estimated density map; e) Apple fruit image with majority of space occupied by background objects; and f) Illustration of successful (yellow rectangle) and unsuccessful (red rectangle) cases of the proposed approach in apple fruit dataset density estimation.}
    \label{fig:Chap5_successfailures}
\end{figure*}
Table \ref{tab:Chap5_psnrSsimAllData} shows the quantitative comparison of the generated density maps for the apple flower and fruit dataset. Our proposed approach showed competitive performance, often outperforming other approaches with higher PSNR (except SCAR for Apple Fruit Dataset) and SSIM values inferring a greater similarity between the ground truth and generated density maps. The generated density maps were 93.8\% and 91.0\% similar to the ground truth density map in terms of luminance, contrast, and structure for the flower and fruit datasets, respectively. Fig. \ref{fig:Chap5_allQualitative} shows a representative density maps for qualitative evaluation.  The generated density maps from the proposed approach showed better localization of flower and fruit regions. The advantage of the proposed approach is more prominent in the flower images. In spite of the fact that the flowers are densely located in clusters  with a high flower-to-flower occlusion, the individual flower regions from the proposed approach were separated with finer details compared to other approaches. We argue this is because of the way feature maps were processed, especially in the decoder section. First, the skip connection from the encoder to the decoder section allowed the low-level feature transfer to the deeper layers, which might have been lost otherwise due to processing and downsampling. Second, similar to the way feature maps were reduced in half in size at each stage in each ConvNeXt - T encoder, the feature maps were progressively up-sampled, concatenated, and processed in the decoder block with feature size at the current sub-block twice the size of the immediate past sub-block. In the case of CSRNet, the feature maps were directly upsampled by a factor of 8 to get the final density map which could be the reason for the non-smooth density map. Third, the segmentation map was helpful in gatekeeping to enhance the relevant features while setting up boundaries for individual flowers and fruits. Furthermore, the proposed approach showed robustness to suppress background objects (flowers/fruits in the immediate back row) that have a similar appearance to the object of interest. We argue that the result is a combined effect of the attention module and segmentation branch to enhance relevant features while suppressing background features.

Fig. \ref{fig:Chap5_successfailures} shows a more detailed (closeup) view of the generated density maps overlayed on top of the canopy images. The Flower dataset was complex compared to the fruit dataset. Regardless of the maturity and color, all fruits appear similar in appearance, sharing similar feature information. However, feature similarity might not be the case in the flowering stage because of the flowering time. The flowering time substantially varies due to the microclimate that could be generated within the canopy due to the closeness to the ground, light interception, irrigation system, and  biological variability. Hence flowers might be in full bloom in some clusters while the majority of flowers might be unopened or only the \textit{King Flower} (Flower that blooms first within flower cluster) in other clusters. Fig. \ref{fig:Chap5_successfailures}a shows the case where the  \textit{King Flower} was open while the lateral flowers within the cluster were still open. Fig. \ref{fig:Chap5_successfailures}b is the representative case where the majority of flowers were not in bloom. Fig. \ref{fig:Chap5_successfailures}c shows the case where flowers in some canopy sections were in full bloom but not in others. Since the proposed approach relies on point annotation, it showed robustness in extracting the features and locating the flowers irrespective of their blooming stage and variability in appearance. However, it is worth noting that the network also missed some of the visible full bloom flowers near the trunk in Fig. \ref{fig:Chap5_successfailures}c.The majority flowers in the background rows were discarded by the network. However, in a few cases (see Fig. \ref{fig:Chap5_successfailures}d), the background flowers were also present in the estimated density map. Such a case was more prominent in the V-trellised system since canopies were in an inverted pyramid shape, and background canopies were much closer compared to 2D fruiting wall architecture with vertical canopies. 
\begin{figure*}[!h]
    
    \centering
    \includegraphics[width=0.38\textwidth]{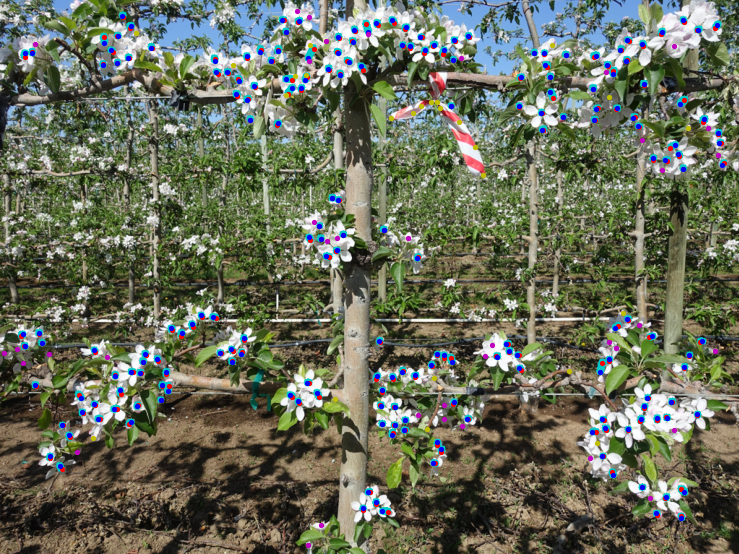}
    \includegraphics[width=0.38\textwidth]{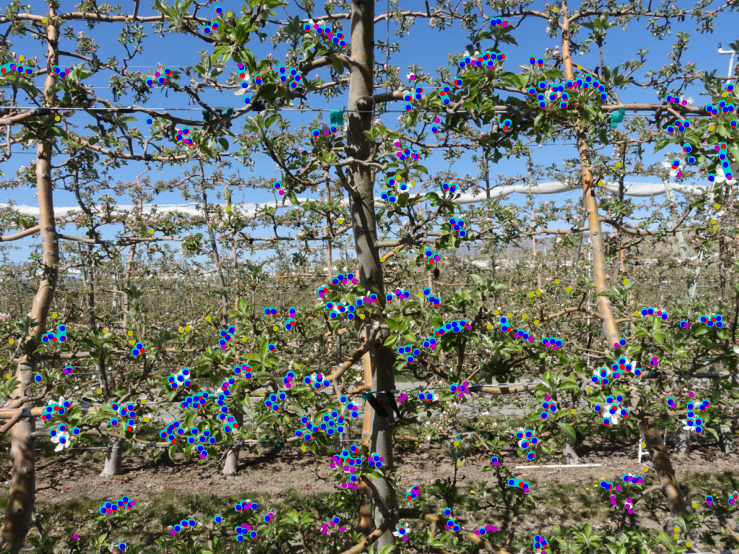} \\
    \vspace{0.8mm}
    \includegraphics[width=0.38\textwidth]{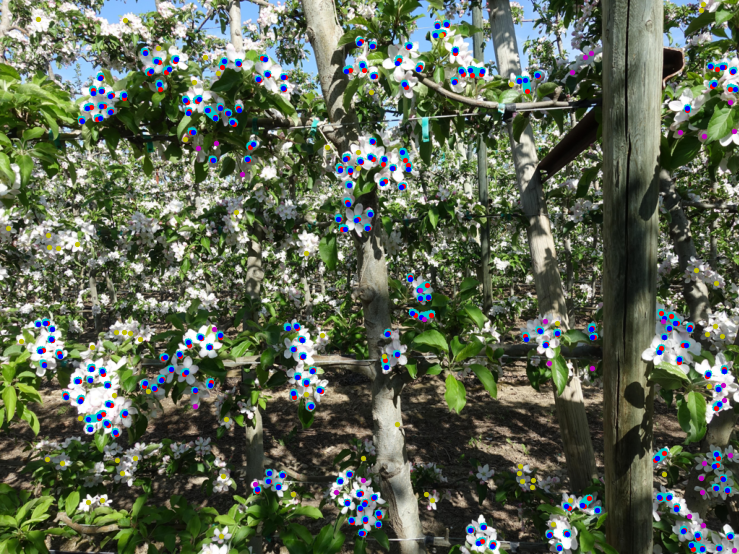}
    \includegraphics[width=0.38\textwidth]{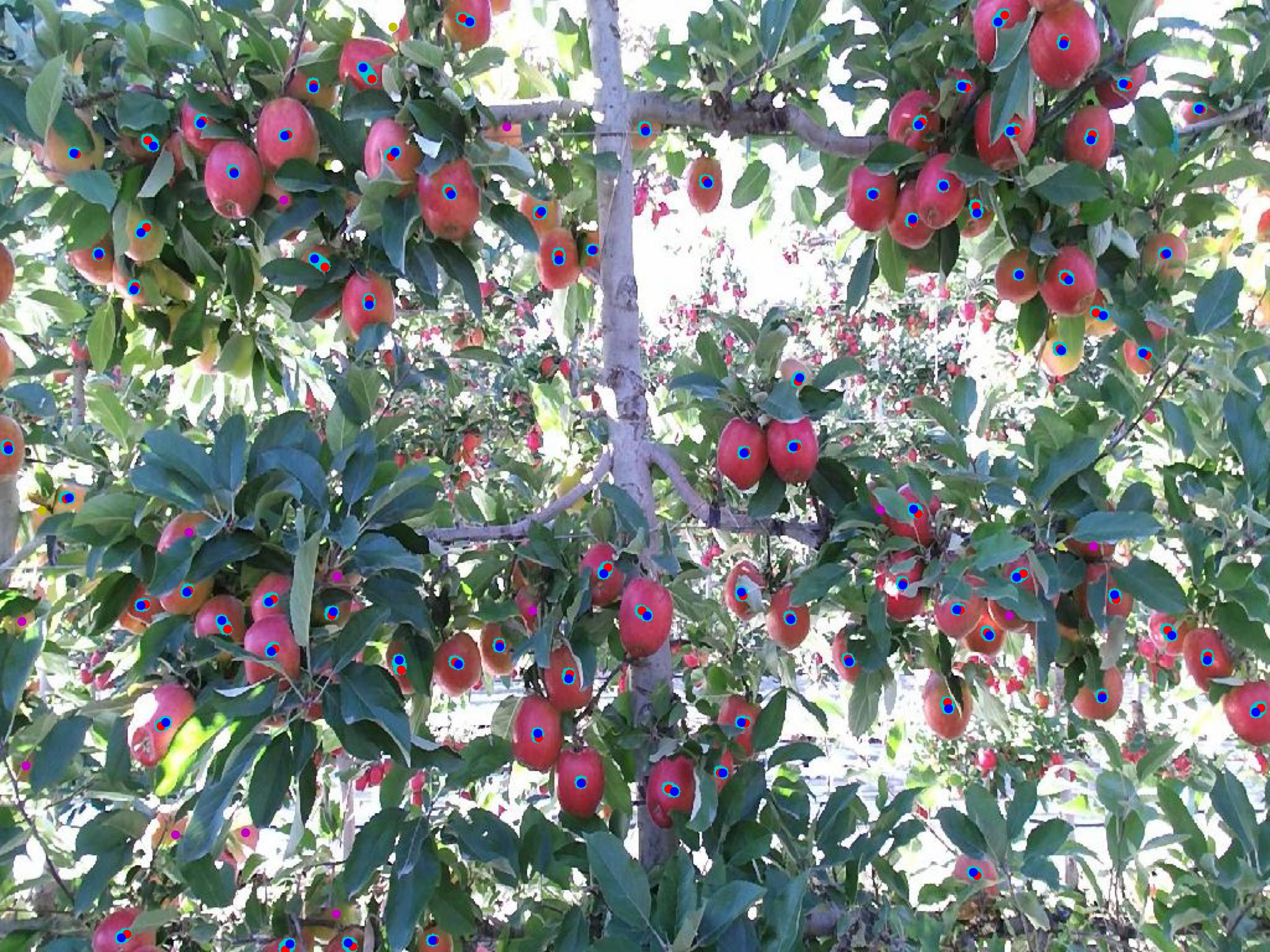}
    \caption{Visualization of flower fruit localization results for flowers and fruits trained in different training architecture. Top-Left: 2D Fruiting Wall (full bloom), Bottom-Left V-trellised (full bloom), Top-Right: V-trellised architecture (unopened flowers), Bottom-Right: 2D Fruiting Wall (Fruits). Color-coded dots represent different ground truths and predicted locations based on their classification designation. Red: ground truth, Blue: True Positive, Magenta: False Negative, and Yellow: False positive. Cyan: Line connecting ground truth to the true positive. }
    \label{fig:Chap5_localization_qualitative}
\end{figure*}

On the other hand, the inclusion of background apples was rarely observed in the fruit dataset. Fig. \ref{fig:Chap5_successfailures}e shows the case where the majority of the image was covered by the background information. Fig. \ref{fig:Chap5_successfailures}f shows the capability and weakness of the regression-based approach. On the left side (yellow rectangle) of Fig. \ref{fig:Chap5_successfailures}f, the apples were hardly visible because of the obstruction caused by the pipe and the branches. Irrespective of the extent of occlusion, the network was able to localize the obstructed apples. The center part of Fig. \ref{fig:Chap5_successfailures}f (red rectangle) shows a case that a side branch divided an apple into three sections. The network mistakenly generated a density map as if three separate apples were present in the canopy. Based on the results in the large number of apple images tested in this work, such cases were few and might have a low impact on overall density estimation and count results as the estimated fruit density map has a 91.0\% similarity to the ground truth map.


\subsection{Flower and Fruit Counting}

The proposed approach showed promising results for density estimation of apples and apple flowers regardless of the flowering stage and appearance, which was also reflected in the count evaluation (Table \ref{tab:Chap5_MseMaeAllData}). The proposed AgRegNet, while being lightweight with the least number of parameters  (9.45M), showed superior performance compared to all evaluated approaches in the complicated dense flower image dataset with the highest PSNR (31.2) and SSIM (0.938) to estimate density map and lowest MAE (18.1) and RMSE (23.8) values to estimate the count. Furthermore, it achieved competitive performance in the fruit image dataset with MAE and RMSE of 3.1 and 4.0, respectively. The percentage MAE $\Big( \%MAE=\frac{MAE}{\# Avg. flower/fruit}*100 \% \Big)$ for the proposed AgRegNet was 5.6\%  for fruit data and 13.7\% for the flower dataset with average fruit and flower per image of 48 and 125 respectively. It is worth noting that remarkably better results were achieved in the fruit dataset compared to the flower dataset. Although fruits were located in clusters, the cluster size was substantially smaller, allowing larger spatial separation between the fruits leading to better-localized density maps and count results. As discussed earlier, the flower dataset was complex because the tightly aligned clusters with flowers in different blooming stages had different feature information.
\begin{table}[!h]
    \centering
    \caption{Comparison of MAE and RMSE of the proposed approach with different flower and fruit counting approaches.}
    \label{tab:Chap5_MseMaeAllData}
   
   \begin{tabular}{lcccc}
        \hline  \hline
        Method &  \multicolumn{2}{l}{Apple Flower} & \multicolumn{2}{c}{Apple Fruit \cite{gao2020multi}} \\
        \cline{2-3} \cline{4-5} 
                        & MAE & RMSE     & MAE & RMSE  \\
        \hline
        CSRNet\cite{Li_2018_csrnet}            &20.9 &29.2 &5.6  &7.1 \\
        SFCN\cite{wang2019sfcn}    &35.8	&52.6	&3.9	&5.1 \\
        SCAR\cite{gao2019scar} &18.4	&25.6	&4.4	&6.1 \\
        Farjon et al. (DRN)\footnote{https://github.com/farjon/Leaf-Counting} \cite{farjon2021leaf} &25.7	&38.0	&\textbf{2.4}	&\textbf{3.2} \\
        AgRegNet (Ours)  &\textbf{18.1} &\textbf{23.8}       &3.1 & 4.0 \\
        \hline \hline
    \end{tabular}
    
\end{table}

\begin{figure*}[!h]
    
    \centering
    \includegraphics[width=0.8\textwidth]{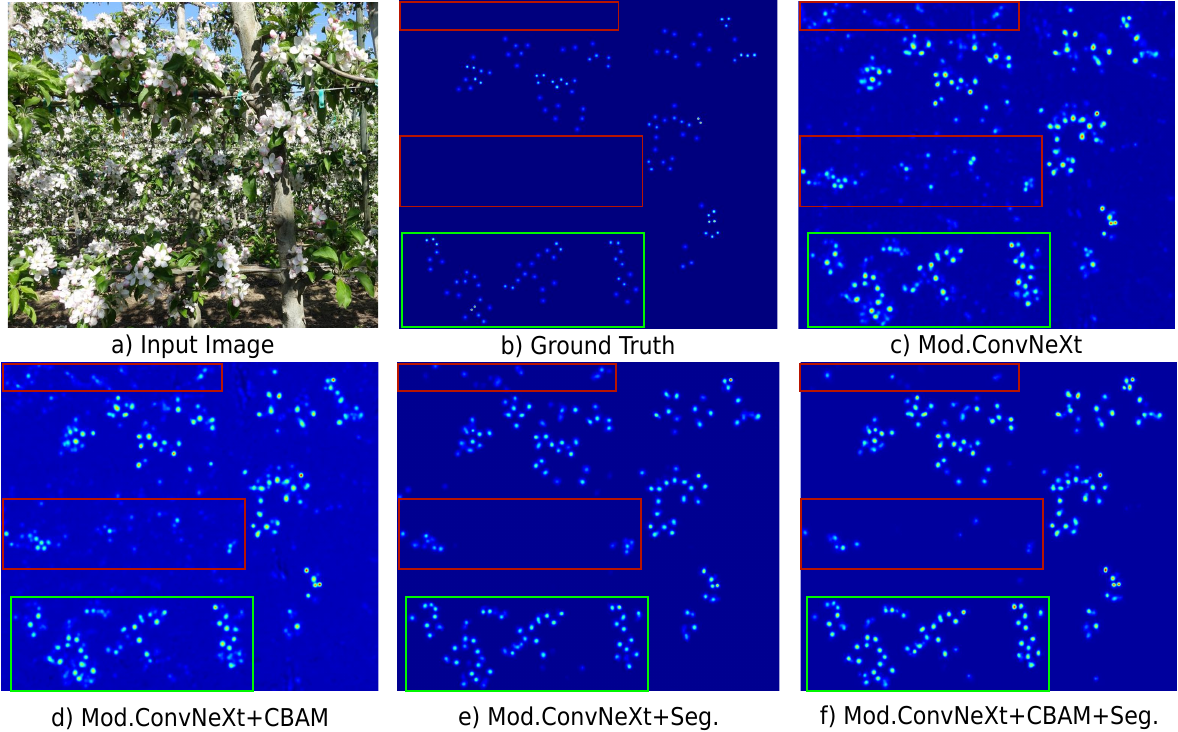}
    \caption{Ablation experiments of the proposed AgRegNet to evaluate the contribution of each module in density map and count estimation. The addition of CBAM and the segmentation branch helped improve the density map estimation and counting accuracy.}
    \label{fig:Chap5_ablation}
\end{figure*}
\subsection{Flower and Fruit Localization} In order to localize the flower and fruit centroid, the density maps were post-processed as per Algorithm \ref{Chap5_localizationalgorithm} (see Section III C). The one-to-one matching between the ground truth and the predicted flower and fruit centroids was analyzed using the mean Average Precision (mAP) and mean Average Recall (mAR) metrics. The mean Average Precision (mAP) and mean Average Recall (mAR) metrics was obtained by averaging the AP and AR values of each image computed by varying the cutoff threshold $\sigma_f$ to $2.2 \sigma_f$ with an increment of 1 (see Section IV c). The cutoff threshold played a role analogous to the Intersection over Union (IoU) threshold in a detection-based approach.  The mAP and mAR values for the fruit dataset were greater than the same for the flower dataset indicating better fruit localization capability with fewer false positive and false negative locations. Numerically, the mAP and mAR values for the fruit dataset were 0.93 and 0.89 respectively, while the same were 0.81 and 0.79 for the flower dataset (see Table \ref{tab:Chap5_localization_results}). However, it is worth noting that the localization result depended on the generated density map, and the post-processing algorithms used to detect peak and one-to-one matching of the estimated centroids.  Fig. 7 visualizes the localization results. The results were promising as the majority of the localized flower and fruit centroids from the proposed approach were correctly associated with the ground truth centroids in images acquired from a commercial orchard.
\begin{table}[!h]
    \centering
    \caption{Mean Average Precision (mAP) and Average Recall (mAR) achieved by the proposed approach in flower and fruit localization}
    \label{tab:Chap5_localization_results}
   \begin{tabular}{lcccc}
        \hline  \hline
        Method &  \multicolumn{2}{l}{Apple Flower} & \multicolumn{2}{c}{Apple Fruit \cite{gao2020multi}} \\
        \cline{2-3} \cline{4-5} 
                        & mAP & mAR     & mAP & mAR    \\
        \hline
        AgRegNet (Ours)       &0.81 & 0.79      &0.93 &0.89      \\

        \hline \hline
    \end{tabular}
    
\end{table}

\subsection{Ablation Experiments}
To investigate the effect of different modules on the performance of the proposed network model, ablation studies were conducted in four different configurations with the Apple Flower dataset. The four configurations used were:
\begin{itemize}
    \item Mod.ConvNeXt-T: Modified ConvNext-T with a U shape and skip connection from the encoder to the decoder section.
    \item Mod.ConvNeXt-T+CBAM: Modified ConvNeXt-T with CBAM attention connected in skip connection within each stage and skip connection from the encoder to decoder.
    \item Mod.ConvNeXt-T+Seg.: Modified ConvNeXt-T with segmentation branch as output in addition to density estimation branch.
    \item Mod.ConvNeXt-T+CBAM+Seg.: The proposed AgRegNet model.
\end{itemize}
\begin{table}[!t]
    \centering
    \caption{Ablation experiment of the proposed approach to evaluate the contribution of each module in density map and count estimation.}
    \label{tab:Chap5_ablationexperiment}
    \scalebox{0.95}{
    \begin{tabular}{lcccc}
         \hline \hline
         Method &MAE &MSE &PSNR &SSIM \\
         \hline 
         Mod.ConvNeXt-T  &27.7	&37.2	&29.8	&0.841 \\
         Mod.ConvNeXt-T+CBAM &25.4	&36.3	&30.0	&0.855 \\
         Mod.ConvNeXt-T+Seg. &20.2	&29.3	&31.1	&0.931 \\
         Mod.ConvNeXt-T+CBAM+Seg. &18.1	&23.8	&31.2	&0.938 \\
         \hline \hline
    \end{tabular}}
    
\end{table}

Table \ref{tab:Chap5_ablationexperiment} shows that the addition of the CBAM and segmentation module showed improved performance of modified ConvNeXt - T. Fig. \ref{fig:Chap5_ablation} shows the progressive refinement of generated density map with the addition of different modules. The modified ConvNeXt - T module suffered from high interference from the background objects, which was reduced to some extent by CBAM. The SSIM value of the modified ConvNeXt - T was 0.841, inferring 84.1\% similarity between the ground truth and the predicted density map.  It was also observed that better density map estimation inferred better localization capability, but it did not necessarily mean increased counting accuracy. Both CBAM and segmentation branches helped enhance the relevant features while suppressing the irrelevant background features. However, quantitative result showed that the effect of the segmentation branch in improving the density map and count was substantially higher compared to CBAM. The addition of the segmentation branch reduced the MAE and RMSE count by 27.1\% and 21.2\% while improving the density map SSIM by 10.7\% compared to the Mod. ConvNeXt - T. On the other hand, the addition of CBAM reduced the MAE and RMSE by 8.3\% and 2.4\% while improving the density map SSIM by 1.6\% compared to the Mod. ConvNeXt - T. As shown in Table \ref{tab:Chap5_ablationexperiment}, the addition of the CBAM module in Mod. ConvNeXt - T with segmentation branch improved the system performance by a nominal value.

\subsection{Practical Applications}

With the simpler annotation technique supporting the generation of a larger training dataset in different orchard environments, and a lighter model structure, the proposed approach is expected to be more robust and practically applicable to various orchard operations. The proposed approach, when used for fruit counting,  could be employed for crop-load estimation and yield estimation in commercial orchards. Fruit count provides yield estimation information allowing efficient and timely management of harvesting resources and developing harvesting strategies. Flower counting is also considered one of the earliest markers of crop yield in a given season. Flower density estimation, localization, and count information could be used for optimizing the currently existing flower thinning strategies using chemical and mechanical thinning approaches for targeted flower thinning applications. Furthermore, for robotic blossom thinning, flower cluster segmentation and flower count information can be combined to perform targeted selective flower thinning \textit{en masse}  such that a proportion of flower could be removed. Furthermore, the proposed approach offers simpler and more effective flower density estimation to optimize chemical thinning by localizing and counting individual flowers instead of approaches involving cluster detection \cite{farjon2019detection} and segmentation \cite{wang2020side}.  \\

Different studies have been reported for flower cluster segmentation in apples \cite{dias2018apple,dias2018multispecies,farjon2019detection,bhattarai2020automatic}. Additionally, it would be more practical to segment individual flower clusters and estimate the number of flowers within each cluster to thin excess flowers \textit{en masse} instead of removing individual flowers as removal of individual flowers would be time-consuming and highly complicated for a robotic system to operate in a constrained working space. As mentioned before, the proposed AgRegNet is lightweight, with a small number of trainable parameters (9.45M), and is suitable for evolving but still limited image datasets prevalent in agriculture. Because of its small size, AgRegNet could be employed in light computation devices such as cell phones. Other compared approaches have a large number of training parameters (e.g. CSRNet - 16.26M by \cite{Li_2018_csrnet}, and DRN - 27.32M by \cite{farjon2021leaf}). Furthermore, the inference time of the proposed approach is substantially lower (14.2 milliseconds) compared to Mask R-CNN (238 milliseconds) employed in the same system for flower cluster detection and segmentation \cite{bhattarai2020automatic}, which improves practical adoption of the proposed model to real-time field operations.

It is noted that the images used in this work were single front view shots which might not display all the flowers/fruits present in the canopy due to the occlusion and limited field of view of the camera. Therefore, the robustness of the proposed approach could be further improved by leveraging multi-view images including the images from both sides of the canopy, using illumination invariant active lighting system \cite{silwal2021robust}, and eliminating unnecessary background objects via depth filtering. We also expect that increasing the number of annotated images will benefit the model for density map estimation and counting in varying orchard conditions and structures.
 
\section{Conclusion}
In this article, a segmentation-assisted regression-based approach for flower and fruit density estimation, counting, and localization in an unstructured orchard environment was presented. Experiments were conducted in the apple flower and fruit datasets to evaluate the network accuracy. Through this study, the following conclusions were drawn. 

\begin{itemize}
    \item Object density estimation, localization, and counting in agricultural fields can be simplified using a regression-based deep learning approach with point annotations in RGB images without explicit detection. The proposed approach correctly localized the majority of the flower (mAP=0.81, mAR=0.79) and fruit (mAP=0.93, mAR=0.89) centroids without precise object detection based on bounding box or polygon annotation.
    \item Proposed AgRegNet showed superior performance in generating well-localized density maps compared to CSRNet, SFCN, and SCAR with high SSIM for flower (SSIM=0.94) and fruit (SSIM=0.91) datasets. For apple flower counting, the proposed approach surpassed CSRNet, SFCN, SCAR and DRN with the lowest MAE of 18.1 (percentage Mean Absolute Error=13.7\%) and RMSE of 23.8. While for the fruit dataset, the proposed approach showed competitive performance with MAE of 3.1 (percentage Mean Absolute Error=5.6\%) and RMSE of 4.0 against DRN, performing best with  MAE of 2.4 and  RMSE of 3.2.
    \item Incorporation of the segmentation branch and attention mechanism was instrumental in improving the robustness to discard background objects and improving the accuracy. With the addition of a segmentation branch the MAE and RMSE of Mod. ConvNeXt - T reduced respectively by 27.1\% and 21.2\% while improving the SSIM of the density map by 10.7\%. Even though the addition of CBAM did not enhance the outcome of the Mod. ConvNeXt - T compared to the same with the segmentation branch, the performance improvement was beneficial.   
\end{itemize}

The results showed that the proposed technique is promising with direct practical applications for growers and is expected to be an essential step towards automating and streamlining flower thinning, crop-load estimation, and yield prediction in the specialty crop production industry. We will continue to work on improving the generalizability of the proposed method by further improving it algorithmically and by including images from multiple canopy views and wider varieties of fruit and vegetables. 
\ifCLASSOPTIONcaptionsoff
  \newpage
\fi

\bibliographystyle{IEEEtran}

\end{document}